\documentclass{article}

 \PassOptionsToPackage{numbers, sort&compress}{natbib}
 
\usepackage[final]{neurips_data_2021}

\usepackage[utf8]{inputenc} 
\usepackage[T1]{fontenc}    
\usepackage[hidelinks]{hyperref} 
\usepackage{url}            
\usepackage{booktabs}       
\usepackage{amsfonts}       
\usepackage{nicefrac}       
\usepackage{microtype}      
\usepackage{xcolor}         

\usepackage{subcaption}
\usepackage{amsmath,amssymb}
\usepackage{graphicx}

\usepackage{multibib} 
\newcites{si}{Additional References for the Appendix} 
\newcommand{\beginsupplement}{ 
\setcounter{section}{0}
\renewcommand{\thesection}{S\arabic{section}} %
\renewcommand{\thesubsection}{\thesection.\arabic{subsection}}
\setcounter{table}{0}
\renewcommand{\thetable}{S\arabic{table}} %
\setcounter{figure}{0}
\renewcommand{\thefigure}{S\arabic{figure}} %
}

\def\itemspaces{-0.9em}
\def\preparagraphspaces{-0.5em}

\newcommand{\gray}[1]{\textcolor{lightgray}{#1}}
\newcommand{\showcomments}{no}
\newcommand{\symfootnote}[1]{%
\let\oldthefootnote=\thefootnote%
\stepcounter{mpfootnote}%
\addtocounter{footnote}{-1}%
\renewcommand{\thefootnote}{\fnsymbol{mpfootnote}}%
\footnote{#1}%
\let\thefootnote=\oldthefootnote%
}

\usepackage{ifthen}

\newcommand\mli[1]{
    \ifthenelse{\equal{\showcomments}{yes}}{{\color{red}[Mu: #1]}}{\ignorespaces}
}

\makeatletter
\newcommand{\printfnsymbol}[1]{%
  \textsuperscript{\@fnsymbol{#1}}%
}
\makeatother

\def\benchlink{\url{https://github.com/sxjscience/automl_multimodal_benchmark}}
\def\gitlink{\url{https://github.com/sxjscience/automl_multimodal_benchmark}}
\def\tutorialink{\url{https://auto.gluon.ai/stable/tutorials/tabular_prediction/tabular-multimodal-text-others.html}}

\def\numdata{18} 

\def\preembed{Pre-Embedding}
\def\textembed{Text-Embedding}
\def\multimodalembed{Multimodal-Embedding}
\def\mmlinearensemble{Weighted-Ensemble}
\def\mmstackensemble{Stack-Ensemble}
\def\mmalltext{All-Text}
\def\mmfuseearly{Fuse-Early}
\def\mmfuselate{Fuse-Late}

\def\agweighted{AG-Weighted}
\def\agstack{AG-Stack}
\def\ngram{N-Gram}

\def\hto{H2O AutoML}
\def\htow2v{\hto{} + Word2Vec}
\def\htoemb{\hto{} + \preembed}

\def\roberta{RoBERTa}
\def\electra{ELECTRA}

\def\textnet{Text-Net}
\def\multimodalnet{Multimodal-Net}

\title{Benchmarking Multimodal AutoML \\ for Tabular Data with Text Fields}

\author{%
  Xingjian Shi\thanks{Equal contribution.} \hfill 
  \texttt{xjshi@amazon.com} 
  \\
  \textbf{Jonas Mueller}\footnotemark[1] \hfill 
  \texttt{jonasmue@amazon.com} 
  \\
  \textbf{Nick Erickson} \hfill 
  \texttt{neerick@amazon.com} 
  \\
  \textbf{Mu Li} \hfill 
  \texttt{mli@amazon.com}
  \\
  \textbf{Alexander J. Smola} \hspace*{10mm} \hfill  
  \texttt{alex@smola.org} \\[0.2em]
    Amazon Web Services
}

\begin{document}

\maketitle

\begin{abstract}
We consider the use of automated supervised learning systems for data tables that not only contain numeric/categorical columns, but one or more text fields as well. 
Here we assemble \numdata{} multimodal data tables that each contain some text fields and stem from a real business application. 
Our publicly-available benchmark\footnote{Benchmark is available at: \benchlink{}} enables researchers to comprehensively evaluate their own methods for supervised learning with numeric, categorical, and text features. 
To ensure that any single modeling strategy which performs well over all \numdata{} datasets will serve as a practical foundation for multimodal text/tabular AutoML, the diverse datasets in our benchmark vary greatly in: 
sample size, problem types (a mix of classification and regression tasks), number of features (with the number of text columns ranging from 1 to 28 between datasets), as well as how the predictive signal is decomposed between text vs.\ numeric/categorical features (and predictive interactions thereof). 
Over this benchmark, we evaluate various straightforward pipelines to model such data, including standard two-stage approaches where NLP is used to featurize the text such that AutoML for tabular data can then be applied. 
Compared with human data science teams, the fully automated methodology\footnote{Open-source available to easily run on your own data: \url{https://github.com/awslabs/autogluon}} that performed best on our benchmark (stack ensembling a multimodal Transformer with various tree models) also manages to rank 1st place when fit to the raw text/tabular data in two MachineHack prediction competitions and 2nd place (out of 2380 teams) in Kaggle's Mercari Price Suggestion Challenge.
\end{abstract}

\section{Introduction}
Despite recent data proliferation, the practical value of machine learning (ML) remains hampered by an inability to quickly translate raw data into accurate predictions. 
Automatic Machine Learning (AutoML) aims to address this via pipelines that can ingest raw data, train models, and output accurate predictions, all without human intervention \cite{automlbook}. 
Given their immense potential, many AutoML systems exist for data structured in tables, which are ubiquitous across science/industry \citep{he2019automl, truong2019towards,  gijsbers2019open}. 

Many data tables contain not only numeric and categorical fields (together referred to as \emph{tabular} here), but also fields with free-form text.  
For example, Table \ref{tab:example-of-real-data} depicts actual data from the website Kickstarter. 
These contain multiple text fields such as the title and description of each funding proposal, numerical fields like the goal amount of funding and when the proposal was created, as well as  categorical fields like the funding currency or country. 
This paper considers tables of this form where rows contain IID training examples (each with a single numeric/categorical value to predict, i.e.\ regression/classification) and the columns used as predictive \emph{features} can contain text, numeric, or categorical values. We refer to the value in a particular row and column as a \emph{field}, where a single text field may actually contain a long text passage (e.g.\ a multi-paragraph item description). 
Despite their potential commercial value, there are currently few (automated) solutions for machine learning with this sort of data that jointly contain numeric/categorical and text features, which we here refer to as \emph{multimodal} or \emph{text/tabular} data. 
Applying existing AutoML tools to such data requires either manually featurizing text fields into tabular format \cite{blohm2020leveraging, h2onlp}, or ignoring the text. Alternatively, one can model just the text with existing natural language processing (NLP) tools. \cite{raffel2020t5,guo2020gluoncv,huggingface,gcpnlp,blazingtext}.

This paper provides foundational tools aiming to spur a practical line of research that evaluates fundamental design choices for automated supervised learning with multimodal datasets that jointly contain text, numeric, and categorical features. 
Even though text commonly appears along with numeric/categorical fields in enterprise data tables, how to model such multimodal data has not been well studied in the literature.  
This stems from a lack of public benchmarks, 
as well as 
existing beliefs that basic featurization of the text should suffice for tabular models to exhibit strong performance \citep{eisenstein2018natural, h2onlp}. 
Here we introduce a new benchmark of \numdata{} multimodal text/tabular datasets involving regression/classification tasks related to real business applications (Section \ref{sec:appdx-dataset-details}), and provide a first systematic evaluation of some generic strategies for supervised learning with such data (Section \ref{sec:experiments}). 

Note that we write \emph{AutoML} to describe any single modeling strategy that remains robustly performant across a diverse set of datasets without manual adjustments. The construction of an effective AutoML system critically relies on having an empirical benchmark of diverse datasets that are representative of real applications the system will be subsequently used for (in order to ensure the system performs well on the right types of data and not only on  certain limited types of data). The experiments over our benchmark presented in this paper merely entail a preliminary evaluation of various straightforward (automated) multimodal modeling strategies that today's data scientists might consider for supervised learning with text/tabular data. Among other discoveries, our benchmark reveals that the conventional strategy of neural embeddings to featurize text for tabular models can be  outperformed by simple alternatives.  
The strategy found to perform best in our benchmark (stack ensembling of tabular models with a multimodal Transformer network) should serve as a strong foundation for multimodal text/tabular AutoML\footnote{A tutorial to easily run this method on your own text/tabular data is provided at: \tutorialink{}}, whose efficacy was subsequently verified in a few data science competitions (demonstrating that our benchmark and analysis have led to important new insights). That said,  much further research is needed in this area, and we hope the public benchmark and open-source tooling introduced here will facilitate practical advances in important text/tabular modeling applications. 

\begin{table}[t!]
\vspace*{-.8em}
    \centering
    \includegraphics[width=0.75\textwidth]{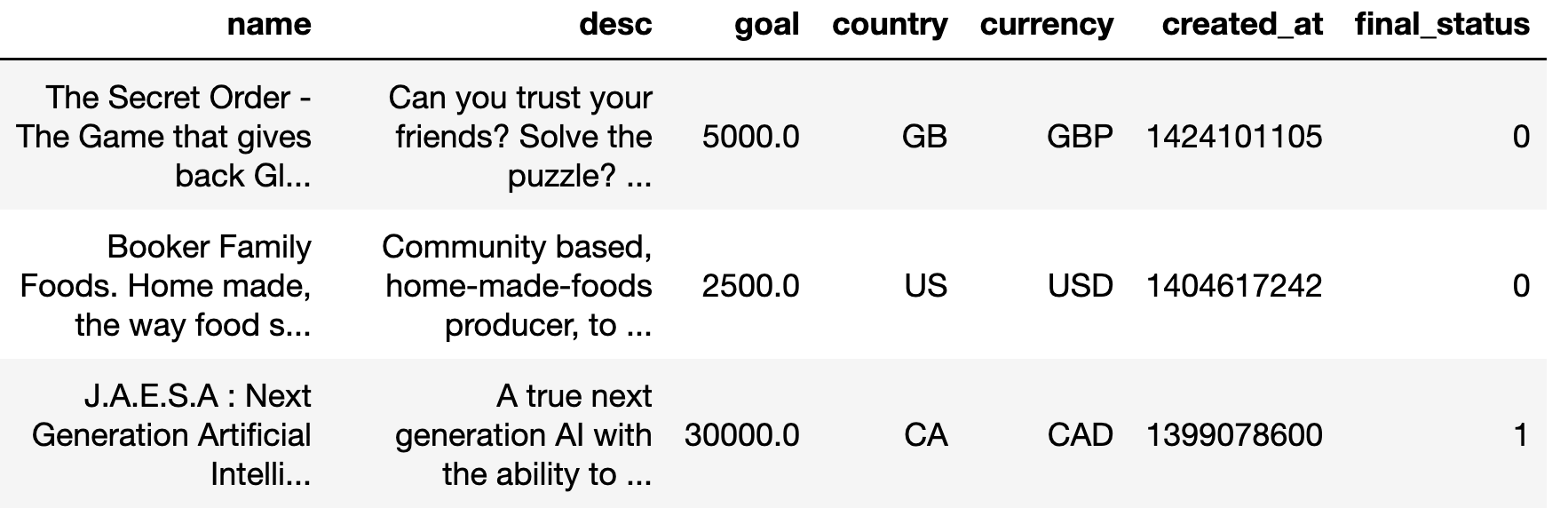}
    \vspace*{0.6em}
     \caption{Example of data in our multimodal benchmark with  text  (\emph{name}, \emph{desc}), numeric (\emph{goal}, \emph{created\_at}), and categorical  (\emph{country}, \emph{currency}) columns. From these features, we want to predict if a Kickstarter project will reach its funding goal or not (\emph{final\_status}).  
     }
     \label{tab:example-of-real-data}
\vspace*{-2em}
\end{table}

\section{Related Work}
\label{sec:relwork}

Many ML courses teach data as vectors in $\mathbb{R}^d$, which is not the case in many practical applications. Thanks to the ubiquity of relevant benchmark data, substantial research has been conducted for properly handling categorical features in a unified manner that generalizes across datasets without sacrificing accuracy \cite{dorogush2018catboost, guo2016entity,larionov2020sampling}. 
Given the prevalence of tabular data composed of  numeric/categorical features, automating the ML process for such data has been the subject of extensive inquiry with major practical impact  \cite{automlbook, he2019automl,  gijsbers2019open}. 
We hope our benchmark spurs similar progress on how to effectively handle text features in tabular data.  

Today, tools for automated learning with text data remain scarce (e.g.\ this dearth forced  \citet{blohm2020leveraging} to turn to tabular AutoML tools for automated text prediction). Instead modern NLP applications primarily require experts who mostly favor 
Transformer networks as their model of choice for text \cite{devlin2019bert, qiu2020pre, raffel2020t5}. 
However existing methods to input numeric/categorical features into Transformers remain  rudimentary \cite{raffel2020t5} and fail to outperform the best tree models for tabular prediction~\cite{huang2020tabtransformer}. 
While multimodal text/tabular Transformer models have been utilized for \emph{table understanding} tasks such as: semantic parsing of facts, cell filling, or relation extraction \cite{yin2020tabert,deng2020turl}, how to best adapt these models for standard classification/regression tasks with text/tabular features remains unstudied to our knowledge (a question that our benchmark might help answer). 
The use of tabular models together with Transformer-like text architectures has  received limited attention \cite{wan2020nbdt, ke2019deepgbm}, and it remains unclear how to optimally leverage their complementary strengths for multimodal data (due to lack of benchmarks). In contrast, a number of entirely-neural architectures have been proposed for multimodal settings~\cite{jin2019auto,wang2020vd,shan2016deep,widedeep}. However the vast majority of these are for $\{$image, text$\}$ data~\cite{sun2019videobert,singh2020mmf,audebert2019multimodal,radford2021learning}, but the gap between neural networks and alternative models is far greater for images than for tabular data \cite{huang2020tabtransformer}. In short, it remains unclear what are the best generic ML pipelines for text/tabular data based on models available today.


Large, sufficiently diverse/representative, public benchmarks have spurred significant progress in tabular AutoML \cite{gijsbers2019open, escalante2020automl, agtabular, zoller2019survey} and NLP \cite{wang2019superglue,nie2020adversarial,gehrmann2021gem,lakew2020low}. 
However we are not aware of any analogous benchmarks for evaluating multimodal text/tabular ML. There do exist a few miscellaneous text/tabular datasets scattered throughout popular ML data repositories \cite{asuncion2007uci,OpenML2013}, but these are mostly small academic datasets that are not representative of modern applications with significant practical value. In contrast, multiple prediction competitions each involving a single real-world text/tabular dataset have been held, but winning solutions have heavily relied on dataset/domain-specific tricks of limited generalizability (c.f.\ mercari \cite{mercariwin} and jigsaw dataset descriptions in Section \ref{sec:datasetdescriptions}). 
Here we aggregate multimodal datasets from competitions and other industry sources into one benchmark that aims to reveal unifying principles for powerful generic modeling of this form of data.

\section{A Benchmark for Supervised Learning with Text/Tabular Data}
\label{sec:appdx-dataset-details}

\begin{table*}[!b]
    \vspace*{-1em}
    \centering
    \resizebox{0.95\linewidth}{!}{%
    \begin{tabular}{llllllllll}
        \toprule
        \textbf{Dataset ID} & \textbf{\#Train} & \textbf{\#Test} & \textbf{\#Cat.} & \textbf{\#Num.}  & \textbf{\#Text} & \textbf{Task} & \textbf{Metric} & \textbf{Prediction Target} \\
        \midrule
               \href{https://machinehack.com/hackathons/product_sentiment_classification_weekend_hackathon_19/overview}{prod} & 5,091 & 1,273 & 1 & 0 & 1 &  multiclass & accuracy &  sentiment associated with product review \\
        \href{https://machinehack.com/hackathons/predict_the_data_scientists_salary_in_india_hackathon/overview}{salary} & 15,841 & 3,961 & 1 & 0 & 5 & multiclass & accuracy & salary range in data scientist job listings
        \\
        \href{https://www.kaggle.com/tylerx/melbourne-airbnb-open-data}{airbnb} & 18,316 & 4,579 & 37 & 24 & 28  &  multiclass & accuracy & price label of Airbnb listing \\
        \href{https://archive.ics.uci.edu/ml/datasets/online+news+popularity}{channel} & 20,284 & 5,071 & 1 & 15 & 1 & multiclass & accuracy  &  news category to which article belongs \\
        \href{https://www.kaggle.com/zynicide/wine-reviews}{wine} & 84,123 & 21,031 & 0 & 2 & 3 & multiclass & accuracy & which variety of wine (type of grape)
        \\\hline
        \href{https://www.kaggle.com/PromptCloudHQ/imdb-data}{imdb} & 800 & 200 & 0 & 7 & 4  & binary & roc-auc & whether film is a drama  \\
        \href{https://www.kaggle.com/shivamb/real-or-fake-fake-jobposting-prediction}{fake} & 12,725 & 3,182 & 2 & 0 & 3 & binary & roc-auc & whether job postings are fake  \\
        \href{https://www.kaggle.com/codename007/funding-successful-projects}{kick} & 86,502 & 21,626 & 3 & 3 & 3 & binary & roc-auc & whether proposed Kickstarter project will achieve funding goal \\
        \href{https://www.kaggle.com/c/jigsaw-unintended-bias-in-toxicity-classification}{jigsaw} & 100,000 & 25,000 & 2 & 27 & 1 & binary & roc-auc  & whether social media comments are toxic  \\\hline
        \href{https://www.kaggle.com/c/google-quest-challenge}{qaa} & 4,863 & 1,216 & 1 & 0 & 3 & regression & $R^2$ & subjective type of answer (in relation to question) \\
        \href{https://www.kaggle.com/c/google-quest-challenge}{qaq} & 4,863 & 1,216 & 1 & 0 & 3 & regression & $R^2$  & subjective type of question (in relation to answer) \\
        \href{https://machinehack.com/hackathons/predict_the_price_of_books/overview}{book} & 4,989 & 1,248 & 1 & 2 & 5 & regression & $R^2$  & price of books \\
        \href{https://www.kaggle.com/PromptCloudHQ/all-jc-penny-products}{jc} & 10,860 & 2,715 & 0 & 2 & 3 & regression & $R^2$ & price of JC Penney products on their website \\
        \href{https://www.kaggle.com/nicapotato/womens-ecommerce-clothing-reviews}{cloth} & 18,788 & 4,698 & 2 & 1 & 3 & regression & $R^2$  & customer review score for clothing item \\
        \href{https://www.kaggle.com/PromptCloudHQ/innerwear-data-from-victorias-secret-and-others}{ae} & 22,662 & 5,666 & 3 & 2 & 6 & regression & $R^2$ &  price of American-Eagle inner-wear items on their website \\
        \href{https://archive.ics.uci.edu/ml/datasets/online+news+popularity}{pop} & 24,007 & 6,002 & 1 & 2 & 1 & regression & $R^2$ & online popularity of news article  \\
        \href{https://www.kaggle.com/c/california-house-prices}{house} & 37,951 & 9,488 & 1 & 18 & 20 & regression & $R^2$ & sale price of houses in California \\
        \href{https://www.kaggle.com/c/mercari-price-suggestion-challenge}{mercari} & 100,000 & 25,000 & 3 & 0 & 6 & regression & $R^2$  & price of Mercari online marketplace products \\
        \bottomrule
    \end{tabular}
    }
    \caption{The \numdata{} multimodal datasets that comprise our benchmark. `\#Cat.', `\#Num.' and `\#Text' count the number of categorical, numeric, and text features in each dataset, and `\#Train' (or `\#Test') count the training (or test) examples. In PDF, click on each Dataset ID for link to original data source. 
    }
    \label{tab:benchmark}
\vspace*{0em}
\end{table*}

In designing practical systems for real-world data tables that often contain text, the empirical performance of our design decisions is what ultimately matters. 
Representative benchmarks comprised of many diverse datasets are critical for proper evaluation of AutoML, whose aim is to reliably produce reasonable accuracy on arbitrary datasets without manual user-tweaking. 
Thus we introduce the first public benchmark 
for evaluating multimodal text/tabular ML, which is comprised of \numdata{} tabular datasets, each containing at least one text field in addition to numeric/categorical columns. 
Our new benchmark is publicly  available, as is the code to reproduce all results presented here.  

Our benchmark strives to represent the types of ML tasks that commonly arise in industry today. 
In creating the benchmark, we aimed to include a mix of classification vs.\ regression tasks and datasets from real applications (as opposed to toy academic settings) that contain a rich mix of text, numeric, and categorical columns. 
Table \ref{tab:benchmark} shows it is comprised of datasets that are quite diverse in terms of: sample-size, problem types, number of features, and type of features. 
11 of the datasets contain more than one text field (with 28 text fields in the airbnb dataset). These text fields greatly vary in the amount of text they contain (e.g.\ short product names vs.\ lengthy product descriptions/reviews). The data (and text vocabulary) stem from a mix of of real-world domains spanning: e-commerce, news, social media, question-answering, and product listings (jobs, projects, films, Airbnb). 
Subsequent accuracy results from Table \ref{tab:main} indicate the \numdata{} prediction problems also vary greatly in terms of both difficulty and how the predictive signal is divided between text/tabular modalities. 
To reflect real-world ML issues, we processed the data minimally (beyond ensuring the features/labels correspond to meaningful prediction tasks without duplicate examples) and thus there are arbitrarily-formatted strings and missing values all throughout. 
Methods/systems that perform well across the diverse set of \numdata{} benchmark datasets are thus likely to provide real-world value for an important class of applications.

Each dataset in our benchmark is provided with a prespecified training/test split (usually 20\% of the original data reserved for test set). Methods are not allowed to access the test set during training, and for validation (model-selection, hyperparameter-tuning, etc.) instead must themselves hold-out some data from the provided training data.  As the choice of training/validation split is a key design decision in AutoML, we leave this flexible for different systems to choose in the learning process. 
To facilitate comparison between the different AutoML pipelines presented in this paper, we always used the same AutoGluon-provided training/validation split, which is stratified based on labels in classification tasks. 
Our use of other AutoML frameworks beyond AutoGluon (e.g.\ H2O) allows each framework to choose their own data splitting scheme. 

The benchmark 
GitHub 
repository contains: (i) methods to easily retrieve the individual datasets and train/test splits, 
(ii) code to run all of the ML strategies studied in this paper and reproduce our results, and (iii) the scripts we used to produce each benchmark dataset from the original data source. 
Common modifications made to original data sources to produce the benchmark dataset versions included: defining a practically meaningful prediction task if there was not one associated with the original dataset, omitting duplicated rows, omitting non-predictive features (e.g.\ user ID) and those that were too correlated with the prediction target (making the benchmark too easy otherwise), applying log-transform to prediction targets that correspond to product prices (log-scale errors are more meaningful in most real pricing applications), and down-sampling overly large datasets (mercari, jigsaw) to ensure the benchmark remains computationally accessible. 
Additional descriptions of each dataset are provided in Appendix  \ref{sec:datasetdescriptions}.


\section{Text/Tabular Modeling Pipelines}

Using our benchmark, we conduct a systematic empirical analysis of various strategies for modeling text/tabular data. 
The strategy that performs best across the benchmark can serve as a promising starting point for automated supervised learning with multimodal data tables that contain text. 
Key choices include what models to use (and for which features), and how to optimally combine different models within an overall supervised learning pipeline. 
Our study considers popular modeling paradigms used by practictioners today, including: 
NLP models to featurize text for tabular models  \cite{blohm2020leveraging,eisenstein2018natural,h2onlp}, ensembling of independently-trained text and tabular models \cite{mercariwin}, or end-to-end learning with neural networks that jointly operate on inputs across text and tabular modalities \cite{jin2019auto,widedeep, raffel2020t5}. 
 Below we merely outline the candidate modeling strategies, Appendix \ref{sec:models} provides full descriptions. 

\paragraph{(Multimodal) Transformer Networks} Given their dominance across NLP, the only models we consider for handling raw text are popular Transformer neural networks which have initially been pretrained in an unsupervised fashion over a massive text corpus \cite{vaswani2017attention, devlin2019bert, raffel2020t5, clark2020electra, liu2019roberta,wang2019superglue}. We investigate how the end-to-end deep learning paradigm can be leveraged for simultaneous text and tabular inputs by extending standard Transformer networks into \emph{multimodal Transformer networks} that jointly operate on both text and tabular features. Three multimodal network variants depicted in Figure \ref{fig:fusion-strategy} are considered: (1)  \emph{\mmalltext{}} -- in which all tabular features are converted to strings and input into the Transformer as text, (2) \emph{\mmfuseearly{}} -- in which dense embedding layers map the tabular features into the same vector space as the embedded text tokens such that self-attention and other Transformer layers can be applied to learn low-level interactions across modalities, (3) \emph{\mmfuselate{}} -- a multi-tower network where one branch is a Transformer network for the text, other branches are multilayer perceptrons (MLP) for the numeric/categorical inputs, and the higher-level vector representations of each branch are pooled into a single multimodal vector representation (near the output layer of the overall network) via concatenation. 

\paragraph{Combining Transformers and Tabular Models} As shown in Figure \ref{fig:aggregation}a, we consider featurizing the text into vector format followed by subsequent application of various tabular models \cite{blohm2020leveraging}. Here the text embedding may stem from a pretrained Transformer network that has not been fine-tuned on our data (\emph{Pre-Embedding}), a Transformer only trained on our text fields alone (\emph{Text-Embedding}), or a multimodal Transformer network trained on both our text and tabular fields (\emph{Multimodal-Embedding}). Note that `tabular models' throughout are those trained on only numeric/categorical features, e.g.\ types of tree-based models. 
We also consider simple weighted ensembles that linearly combine the predictions of our Transformer network and various tabular models (\emph{Weighted-Ensemble}, shown in Figure \ref{fig:aggregation}b) where each model takes as input the modalities it is suited for and is independently trained from the other models \cite{agtabular,caruana2004ensemble}. Finally, stack ensembling is alternatively considered to nonlinearly aggregate predictions from the Transformer and tabular models  (\emph{Stack-Ensemble}, shown in Figure \ref{fig:aggregation}c), where an additional tabular `stacker' model is trained using as its features the predictions output by the  independently-trained Transformer and original tabular models  \cite{agtabular,van2007super}. 

In our study, all tabular (numeric/categorical) modeling is simply done via 
AutoGluon-Tabular, an easy to use open-source tool for automated supervised learning on tabular data \citep{agtabular}. We chose AutoGluon because it has been found to produce highly accurate models for diverse tabular datasets   \citep{yoo2020ensemble,fakoor2020fast,bezrukavnikov2021neophyte,smalldata}.
AutoGluon trains and ensembles a diverse suite of popular models for tabular data, including: Gradient Boosted Decision Trees \cite{ke2017lightgbm,dorogush2018catboost,chen2016}, Extremely Randomized Trees \citep{geurts2006extremely}, and MLP Neural Networks \citep{agtabular}. 
While neural networks are typically favored for unstructured data like text, decision tree ensembles have proven to be one of the most consistently performant models for tabular data \citep{kaggletrends, fakoor2020fast, huang2020tabtransformer}. Thus an effective strategy for text/tabular AutoML may need to appropriately combine the complementary strengths of  Transformers and (tree-based) tabular models. 

\section{Experiments}
\label{sec:experiments}

To keep our study tractable, we adopt a sequential decision making process that decomposes the overall supervised learning pipeline design into three stages: 1) determine the appropriate Transformer backbone and fine-tuning strategy for text data alone (Appendix \ref{sec:transformer}), 2) determine the best way to extend this  Transformer to text and tabular inputs (Appendix \ref{sec:multimodalnn}), and 3) determine the best method to combine the best text and tabular models (Appendix \ref{sec:featurizetext} and \ref{sec:ensembling-methods}). 
At each subsequent stage of the study, we explore modeling choices that are specific to that stage and simply use the best choice found in the empirical comparisons of the options available in previous stages. 

For straightforward comparison, we employ the most commonly used classification/regression evaluation metrics that lie in $[0,1]$ for reasonable predictions, with higher values indicating superior performance. We evaluate regression tasks via the coefficient of determination $R^2$, multiclass classification tasks via accuracy, and binary classification tasks via area under the ROC curve (AUC). 

\vspace*{\preparagraphspaces{}}
\paragraph{Choice of Transformer Backbone}  
Our first decision concerns the Transformer network itself, including what architecture and pretraining objective to employ. 
Existing results may not translate to our setting, since Transformers are typically applied to datasets with at most a couple text fields per training example \citep{wang2018glue,wang2019superglue}. 
Here we choose between the (standard, already pretrained) base version of  RoBERTa~\citep{liu2019roberta} or ELECTRA~\citep{clark2020electra}, two popular backbones used across modern NLP applications. 

We first fine-tune the pretrained Transformer models as our sole predictors,  using only the text features in each dataset.  This helps identify which model is better at handling the types of text in our multimodal datasets. 
During fine-tuning of both of the RoBERTa or ELECTRA networks, we additionally consider two tricks to boost performance: 
1) Exponentially decay the learning rate of the network parameters based on their depth~\citep{sun2019fine}. We use a per-layer learning rate multiplier of $\tau^d$ in which $d$ is the layer depth and $\tau$ is the decay factor (set $= 0.8$ throughout). 
2)  Average the weights of the models loaded from the top-$3$ training checkpoints with the best validation scores~\citep{vaswani2017attention}. 

The first section of Table \ref{tab:main} shows that \electra{} performs better than \roberta{} across the text columns in our benchmark datasets. 
Our exponential decay and checkpoint-averaging tricks further boost performance, with the majority of additional gains produced by exponential decay. 
In subsequent experiments, we thus fix \electra{} fine-tuned with both exponential decay and checkpoint-averaging as the model used to handle text features and call it \emph{\textnet{}}. 

\vspace*{\preparagraphspaces{}}
\paragraph{Best Multimodal Network}
Next, we explore the best way to extend the \emph{\textnet{}} model to operate across numeric/categorical inputs in addition to text fields (among the options in Figure \ref{fig:fusion-strategy}). 

Across our datasets, Table \ref{tab:main} shows that the \emph{\mmfuselate{}} strategy outperforms \emph{\textnet{}} and the alternative \emph{\mmalltext{}}/\emph{\mmfuseearly{}} options for producing predictions from multimodal inputs using a single neural network.
We thus fix this \emph{\mmfuselate{}} model as our \emph{\multimodalnet{}} used in subsequent experiments. 

\vspace*{\preparagraphspaces{}}
\paragraph{Aggregating Transformers and Tabular Models}

Having identified a good neural network architecture for multimodal text/tabular inputs, we now study combinations of such models with classical learning algorithms for tabular data (among the options in Figure \ref{fig:aggregation}). Where not specified, the tabular models are those trained by AutoGluon-Tabular (see Appendix~\ref{sec:ag-model-stacker}). 
The third section of Table \ref{tab:main} illustrates that \emph{\mmstackensemble{}} is overall the best aggregation strategy. Ensembling the predictions of \emph{Multimodal-Net} and tabular models is better than instead using the Transformer for text embedding. 
As expected, \emph{\textembed{}} and \emph{\multimodalembed{}} outperform \emph{\preembed{}}, demonstrating how domain-specific fine-tuning improves the quality of learned embeddings. 
\emph{\multimodalembed{}} performs better than \emph{\textembed{}} on some datasets and similarly across the rest, showing it can be beneficial to use text representations contextualized on numeric/categorical information.

\begin{table*}[tb]
    \centering
    \resizebox{1.0\linewidth}{!}{%
    \begin{tabular}{l|cccccccccccccccccc|cc}
        \toprule
        Method & prod & qaq & qaa & cloth & airbnb & ae & mercari & jigsaw & imdb & fake & kick & jc & wine & pop & channel & salary & book & house & avg $\uparrow$ & mrr $\uparrow$ \\
        \midrule
        \gray{Choosing \textnet{}:} & \multicolumn{17}{c}{NLP Backbones and Fine-tuning Tricks (Section \ref{sec:transformer})} & & &\\
        \roberta & 0.588 & 0.412 & 0.268 & 0.700 & 0.344 & 0.953 & 0.561 & 0.960 & 0.731 & 0.929 & 0.751 & 0.615 & 0.811 & -0.000 & 0.301 & 0.396 & 0.151 & 0.821 & 0.572 & 0.07\\
        \electra & 0.705 & 0.410 & 0.356 & 0.718 & 0.349 & 0.955 & 0.586 & 0.965 & 0.750 & 0.824 & 0.754 & 0.606 & 0.813 & 0.003 & 0.315 & 0.457 & 0.466 & 0.857 & 0.605 & 0.09\\
        + Exponential Decay $\tau=0.8$ & 0.728 & 0.436 & 0.431 & 0.743 & 0.337 & 0.953 & 0.579 & 0.963 & 0.852 & 0.963 & 0.760 & 0.664 & 0.808 & 0.004 & 0.308 & 0.447 & 0.568 & 0.841 & 0.632 & 0.12\\
        + Average 3 $\bigstar$ & 0.729 & 0.451 & 0.432 & 0.746 & 0.350 & 0.954 & 0.581 & 0.965 & 0.858 & 0.961 & 0.766 & 0.656 & 0.807 & 0.004 & 0.307 & 0.445 & 0.571 & 0.841 & 0.635 & 0.14\\
        \midrule
        \gray{Choosing \multimodalnet{}:} & \multicolumn{17}{c}{Fusion Strategy \ (Section \ref{sec:multimodalnn}, Figure \ref{fig:fusion-strategy})} & & & \\
        \mmalltext{} & 0.907 & 0.454 & 0.419 & 0.746 & 0.366 & 0.957 & 0.599 & \textbf{0.967} & 0.840 & 0.967 & \textbf{0.799} & 0.645 & 0.810 & 0.013 & 0.480 & 0.465 & 0.585 & 0.892 & 0.662 & 0.28\\
        \mmfuseearly{} & \textbf{0.913} & 0.441 & 0.418 & 0.745 & 0.377 & 0.953 & 0.596 & \textbf{0.967} & 0.843 & 0.960 & 0.770 & 0.653 & 0.806 & 0.013 & 0.474 & 0.458 & 0.548 & 0.901 & 0.658 & 0.21\\
        \mmfuselate{} $\bigstar$ & 0.907 & 0.449 & \textbf{0.445} & 0.747 & 0.395 & 0.958 & 0.603 & 0.966 & 0.857 & 0.961 & 0.773 & 0.639 & 0.812 & 0.015 & 0.481 & 0.468 & 0.571 & 0.907 & 0.664 & 0.22\\
        \midrule
        \gray{Choosing Aggregation:} & \multicolumn{17}{c}{Multimodal Model Aggregation \ (Sections \ref{sec:featurizetext} and \ref{sec:ensembling-methods}, Figure \ref{fig:aggregation})} & & & \\
        \preembed{} & 0.895 & 0.216 & 0.247 & 0.642 & 0.449 & 0.972 & 0.433 & 0.586 & 0.871 & 0.926 & 0.743 & 0.491 & 0.680 & 0.012 & 0.526 & 0.460 & 0.581 & 0.939 & 0.593 & 0.11\\
        \textembed{} & 0.867 & 0.446 & 0.432 & 0.748 & 0.430 & 0.972 & 0.434 & 0.587 & 0.855 & 0.962 & 0.790 & 0.658 & 0.830 & 0.008 & 0.502 & 0.438 & 0.594 & 0.932 & 0.638 & 0.17\\
        \multimodalembed{} & 0.907 & 0.439 & 0.437 & 0.749 & 0.438 & 0.974 & 0.432 & 0.587 & 0.847 & 0.967 & 0.794 & \textbf{0.683} & 0.829 & 0.007 & 0.517 & 0.451 & 0.595 & 0.934 & 0.644 & 0.25\\
        \mmlinearensemble{} & 0.907 & 0.439 & 0.429 & 0.744 & 0.453 & 0.976 & 0.597 & 0.957 & 0.876 & 0.923 & 0.787 & 0.641 & 0.814 & 0.018 & 0.554 & 0.483 & 0.620 & 0.941 & 0.676 & 0.25\\
        \mmstackensemble{} $\bigstar$ & 0.909 & \textbf{0.456} & 0.438 & \textbf{0.751} & 0.459 & 0.977 & \textbf{0.605} & \textbf{0.967} & \textbf{0.878} & 0.964 & 0.797 & 0.624 & 0.836 & \textbf{0.020} & \textbf{0.556} & 0.496 & \textbf{0.638} & \textbf{0.943} & \textbf{0.684} & \textbf{0.69}\\
        \midrule
        \midrule
        & \multicolumn{17}{c}{Tabular AutoML + Feature Engineering Baselines \ (Section \ref{sec:featurizetext})} & & & \\
        \agweighted & 0.891 & 0.046 & 0.076 & -0.002 & 0.426 & 0.841 & 0.098 & 0.587 & 0.845 & 0.686 & 0.668 & 0.004 & 0.173 & 0.016 & 0.549 & 0.226 & 0.222 & 0.934 & 0.405 & 0.08\\
        \agstack & 0.891 & 0.046 & 0.077 & 0.001 & 0.435 & 0.841 & 0.098 & 0.587 & 0.844 & 0.697 & 0.670 & 0.003 & 0.175 & 0.017 & 0.550 & 0.226 & 0.233 & 0.934 & 0.407 & 0.09\\
        \agweighted + \ngram & 0.892 & 0.426 & 0.382 & 0.610 & 0.450 & 0.978 & 0.526 & 0.909 & 0.842 & 0.966 & 0.772 & 0.357 & 0.829 & 0.019 & 0.546 & 0.484 & 0.591 & 0.941 & 0.640 & 0.17\\
        \agstack + \ngram & 0.895 & 0.414 & 0.383 & 0.654 & \textbf{0.466} & \textbf{0.979} & 0.569 & 0.915 & 0.850 & \textbf{0.968} & 0.775 & 0.612 & \textbf{0.842} & \textbf{0.020} & 0.548  & 0.494 & 0.600 & 0.943 & 0.663 & 0.43\\
        \hto & 0.869 & 0.247 & 0.159 & 0.163 & 0.329 & 0.976 & 0.430 & 0.531 & 0.813 & 0.756 & 0.669 & 0.411 & 0.478 & 0.014 & 0.530 & 0.525 & 0.444 & 0.939 & 0.516 & 0.11\\
        \htow2v & 0.859 & 0.244 & 0.285 & 0.624 & 0.347 & 0.973 & 0.534 & 0.847 & 0.827 & 0.943 & 0.755 & 0.443 & 0.778 & 0.013 & 0.524 & \textbf{0.528} & 0.586 & 0.932 & 0.613 & 0.14\\
        \htoemb & 0.846 & 0.227 & 0.312 & 0.644 & 0.367 & 0.969 & 0.282 & 0.572 & 0.874 & 0.893 & 0.738 & 0.549 & 0.571 & 0.007 & 0.483 & 0.483 & 0.523 & 0.933 & 0.572 & 0.09\\
        \bottomrule
    \end{tabular}
    }
    \vspace*{0em}
    \caption{Accuracy (and $R^2$, AUC) of AutoML strategies over our multimodal benchmark. Column \textbf{avg} lists each method's average score across datasets (i.e.\ how \emph{much} methods differ in overall performance) and \textbf{mrr} its mean reciprocal rank among all evaluated methods (i.e.\ how \emph{often} a method outperforms others). 
    Each subsection encapsulates a design stage ($\bigstar$ marks variant with best avg).
    }
    \label{tab:main}
\vspace*{-1em}
\end{table*}

\vspace*{\preparagraphspaces{}}
\paragraph{AutoGluon Baselines}
As many of our results use the tabular models in AutoGluon \cite{agtabular}, we also compare different variants of AutoGluon-Tabular (without our \emph{\multimodalnet{}}) as baselines:
\\[\itemspaces{}]

\noindent
\emph{\agweighted{}} / \emph{\agstack{}}:  We train AutoGluon with weighted / stack ensembling of its tabular models, here ignoring all text columns. Thus,  baseline ML performance of tabular models without using any text fields can be established via the AG-Weighted/Stack numbers in Table \ref{tab:main} (without N-Gram).
\\[\itemspaces{}]

\noindent 
\emph{\agweighted{} + \ngram{}} / \emph{\agstack{} + \ngram{}}: Similar to \emph{\agweighted{}} / \emph{\agstack{}}, except we first use AutoGluon's \ngram{} featurization \cite{eisenstein2018natural}  to encode all text in tabular form. 

The performance gap between AutoGluon-Tabular with and without \ngram{}s can reveal (an approximate lower bound for) how much extra predictive value is provided by the text features in each dataset. Inspecting these gaps, we find that, compared to the tabular features, text features contain most of the predictive signal in some datasets (qaq, qaa, cloth, mercari, jc), and far less signal in other datasets (prod, imdb, channel), again highlighting the diversity of our benchmark. 
Note that our proposed \emph{\mmstackensemble{}} performs relatively well across all types of datasets, regardless how the predictive signal is allocated  between text and tabular features. 
\\[\itemspaces{}]

\vspace*{\preparagraphspaces{}}
\paragraph{H2O Baselines} 
In addition to AutoGluon, we also run another popular open-source AutoML tool offered in H2O \cite{h2oversion}. 
Since \hto{} is not designed for the text in our multimodal data tables, we try combining H2O's NLP tool~\cite{h2onlp} and tabular AutoML tool~\cite{ledell2020h2o}. 
\\[\itemspaces{}]

\noindent 
\emph{\hto{}}: We run H2O AutoML directly on the original data of our benchmark. As a tabular AutoML framework, H2O AutoML is assumed to ignore text features, but H2O categorizes feature types differently than us and automatically treats some columns we consider to be text as categorical instead. 
\\[\itemspaces{}]

\noindent 
\emph{\htow2v{}}: We run H2O's word2vec algorithm to featurize text fields and then \hto{} on the featurized data, following their  recommended procedure \cite{h2onlp}.
\\[\itemspaces{}]

\noindent 
\emph{\htoemb{}}: We featurize each text field using embeddings from a pretrained \electra{} Transformer, as in \emph{\preembed{}}, followed by H2O AutoML on the featurized data table.
\\[\itemspaces{}]

The last section of Table \ref{tab:main} shows that while these powerful AutoML ensemble predictors  can outperform our individual neural network models (particularly for datasets with more predictive signal in the tabular features), our  \emph{\mmstackensemble{}} and \emph{\mmlinearensemble{}} are superior overall. 

Table \ref{tab:main} shows the accuracy for some datasets significantly improves when models utilize the text features rather than ignoring them. Predictive performance of \emph{AG-stack} (baseline tabular model that ignores text) vs.\ \emph{Stack-Ensemble} (our extension that leverages text) is 0.098 vs.\ 0.605 on mercari, 0.670 vs.\ 0.797 on kick, and 0.175 vs.\ 0.836 on wine. On these datasets, modeling the tabular features brings clear improvements over the text alone given the performance of \emph{Text-Net} (our best text Transformer model that ignores tabular features) is only: 0.581 on mercari, 0.766 on kick, and 0.807 on wine. In certain applications, accuracy improvements of this magnitude may have significant commercial value, thus highlighting the benefits of multimodal modeling of text/tabular data. 

\paragraph{Performance in Real-world ML Competitions}
Some datasets in our multimodal benchmark originally stem from previous ML competitions. For these (and other recent competitions with text/tabular data), we fit the automated strategy that performed best in our benchmark (\emph{\mmstackensemble{}}) to the official competition dataset, without manual adjustment or data processing. We then submit its resulting predictions on the competition test data to be scored, which enables us to see how they fare against the manual efforts of human data science teams. 

The \emph{\mmstackensemble{}} strategy achieves 1st place historical leaderboard rank in two MachineHack prediction competitions: \emph{Product Sentiment Classification}\footnote{ \scriptsize \url{https://www.machinehack.com/hackathons/product_sentiment_classification_weekend_hackathon_19/overview} (``Anonymous Submission ID 1556'' entry)} and \emph{Predict the Data Scientists Salary in India}\footnote{\scriptsize \url{https://machinehack.com/hackathons/predict_the_data_scientists_salary_in_india_hackathon/overview} (``Xingjian Shi'' entry)}, and 2nd place in another: 
\emph{Predict the Price of Books}\footnote{\scriptsize \url{https://machinehack.com/hackathons/predict_the_price_of_books/overview} }. This same strategy also achieves 2nd place on the historical leaderboard of two Kaggle competitions: \emph{California House Prices}\footnote{\scriptsize \url{https://www.kaggle.com/c/california-house-prices} \ (``sxjscience'' entry)} and \emph{Mercari Price Suggestion Challenge}\footnote{\scriptsize 
 \url{https://github.com/sxjscience/automl_multimodal_benchmark/blob/main/competition_submissions/mercari_submission_screenshot.png} }, where the latter was a very popular Kaggle competition in which 2380 teams participated (with \$100,000 prize offered to winner). 
These results show that a straightforward AutoML strategy identified via preliminary analysis of our benchmark is already competitive with data scientists on real-world text/tabular datasets that possess great commercial value. Extensive studies of the benchmark will presumably reveal even more effective strategies.

\paragraph{Feature Importance Analysis}
Feature importance can help us understand what drives a ML system's accuracy and whether text fields in a dataset are worth their overhead. For two representative datasets from our benchmark, we compute \emph{permutation feature importance} \citep{breiman2001random} for our trained models, 
which is defined as the drop in prediction accuracy after values of only this feature (which are entire text fields for a text column) are shuffled in the test data (across rows). We only shuffle original column values so our importance scores are not biased by preprocessing/featurization decisions (except in how these directly affect model accuracy). 

\begin{figure}[tb!]
    \centering
    \begin{subfigure}[b]{0.49\textwidth}
         \centering
         \includegraphics[width=\textwidth]{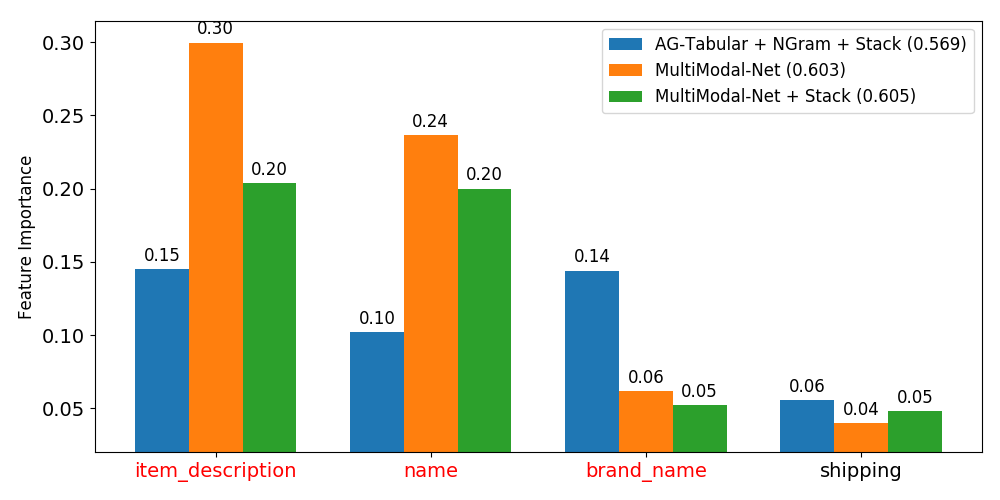}
         \vspace*{-1.6em}
         \caption{Permutation importance in ``mercari''.}
         \label{fig:mercari-feature-importance}
     \end{subfigure}
     \hfill
     \begin{subfigure}[b]{0.49\textwidth}
         \centering
         \includegraphics[width=\textwidth]{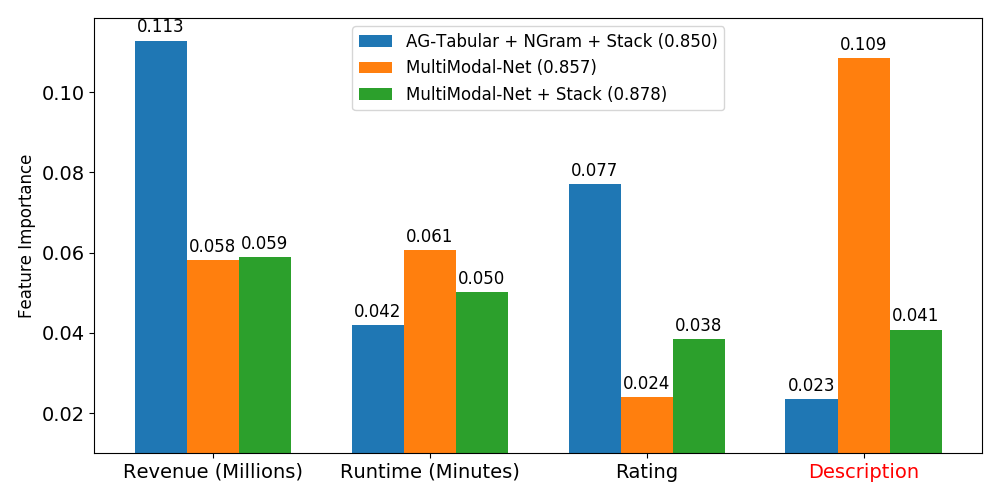}
         \vspace*{-1.6em}
         \caption{Permutation importance in ``imdb''.}
         \label{fig:imdb-feature-importance}
     \end{subfigure}
     \vspace*{-0.1em}
    \caption{Importance of text vs.\ tabular features for three models in two datasets (text features in red). Here \emph{MultiModal-Net + Stack} corresponds to the \emph{Stack-Ensemble} method from Figure \ref{fig:aggregation}c. 
    }
    \label{fig:feature-importance}
\vspace*{-0.3em}
\end{figure}

Figure~\ref{fig:feature-importance} shows that both our \emph{\multimodalnet{}} and \emph{\mmstackensemble{}} containing this network may rely more heavily on text features than the 
\emph{\agstack{}+\ngram{}} 
baseline. 
With more powerful modeling of text fields, models often begin to rely more heavily on the text fields. An exception here is the \emph{brand\_name} feature in mercari, but this feature  usually contains just a single word in its fields. 
Furthermore, the \emph{Multimodal-Net} places less importance on the tabular features, demonstrating how purely neural network approaches are less effective for modeling numeric/categorical data compared to alternative tree-based tabular models. It is thus useful to combine both multimodal-Transformer and tabular models in order to ensure we are most effectively modeling both the text and tabular features.

\section{Discussion}

Lacking public benchmarks, academic research on ML for multimodal text/tabular data has not matched industry demand to derive practical value from such data. This paper provides evidence that generic best practices for such data remain unclear today: we simply evaluated a few basic strategies on our benchmark and found a single automated strategy that turns out to outperform human data scientists in numerous historical prediction competitions involving diverse text/tabular data. 
This strategy uses a stack ensemble (Appendix \ref{sec:ensembling-methods}) of tabular models trained on top of predictions from other tabular models and a \emph{Multimodal-Net} (depicted in Figure \ref{fig:stack-ensemble}). The latter network is based on a \emph{Fuse-Late} architecture (depicted in Figure \ref{fig:fusion-strategy-late-fusion}) with concatenation of text, numeric, and categorical representations (where text representations are produced via the ELECTRA Transformer backbone) and is trained via fine-tuning with exponential learning rate decay and checkpoint averaging. 

Note that this stack ensemble strategy was simply the method that performed best over our benchmark datasets. The fact that this strategy is also highly competitive in numerous text/tabular prediction competitions highlights the utility of our benchmark in revealing performant modeling techniques, indicating that the benchmark is sufficiently diverse and representative of real-world text/tabular prediction tasks. 
Our benchmark analysis challenges certain conventional beliefs: 
\begin{itemize}
\vspace*{-0.5em}
\item Neural embedding of text followed by tabular modeling (\emph{Pre/Text-Embedding})  \cite{blohm2020leveraging, h2onlp} is often outperformed by N-gram featurization (\emph{\agstack{} + \ngram{}}) or leveraging predictions from text neural networks  (\emph{\mmstackensemble{}}) rather than their representations (embeddings). 
Given the success of pretrained Transformers across NLP, we are surprised to find both \ngram{}s and word2vec here provide superior text featurization than \emph{\preembed{}}.
\item In the architecture of 
 multimodal networks for classification/regression, newer ideas to fuse modalities in early layers (i.e.\ \emph{\mmfuseearly{}}/\emph{\mmalltext{}} Transformers with cross-modality attention \cite{singh2020mmf,raffel2020t5,hu2021transformer}) are not necessarily superior to older multi-tower \emph{\mmfuselate{}} architectures that fuse representations in higher layers closer to the output  \cite{jin2019auto,widedeep,audebert2019multimodal}.  
\item An end-to-end multimodal neural network is surpassed by stack ensembling this \emph{\multimodalnet{}} with tabular models trained in separate stages rather than end-to-end (\emph{\mmstackensemble{}}). 
\vspace*{-1.1em}
\end{itemize} 
Previously anticipated conclusions that are empirically validated by our benchmark include:
\begin{itemize}
\vspace*{-0.2em}
\item Text featurization is better via fine-tuned networks  (\emph{\textembed{}}) than pretrained ones (\emph{\preembed{}}), and slightly better via a fine-tuned multimodal network (\emph{\multimodalembed{}}), whose text embeddings benefit from contextualization on the tabular features.
\item Able to exploit predictive interactions between different modalities, stack ensembling outperforms simple weighted ensembling, yet it still facilitates modular system design. 
\vspace*{-0em}
\end{itemize}


Further analysis of our text/tabular benchmark can reveal many more practical ML insights, including under which data conditions certain methods perform better than others. Future research should investigate different data preprocessing pipelines, which are known to play an important role in AutoML.  Other important questions not considered in our preliminary study include \emph{how to best}: Handle many long text fields? Perform multimodal feature selection? Apply feature engineering that combines synergistically with learned neural network representations? 
Allocate limited training/HPO time between cheaper tabular models and more expensive text neural networks? 
We hope our public benchmark spurs the AutoML community to broaden their methods' applicability to more data types. 

\clearpage
\bibliographystyle{abbrvnat}
\bibliography{multimodalautoml}

\clearpage
\beginsupplement
\appendix
\begin{center}
    \Large \textbf{Appendix}
\end{center}

\section{Descriptions of Text/Tabular Modeling Pipelines}
\label{sec:models}

Here we fully describe the various straightforward modeling strategies that we evaluate over our benchmark in order to identify performant baselines for automated supervised learning with multimodal data tables that contain text. 
Recall our study aims to cover popular variants of text/tabular modeling used in practice today, including: NLP models to featurize text for tabular models \citesi{blohm2020leveraging,eisenstein2018natural,h2onlp}, ensembling of independently-trained text and tabular models \citesi{mercariwin}, or end-to-end learning with neural networks that jointly operate on inputs across text and tabular modalities \citesi{jin2019auto,widedeep, raffel2020t5}. 
We first consider the latter paradigm of multimodal neural network models, which in subsequent sections are also considered for text featurization and ensembling with tabular models.

\subsection{Transformer Models for Text}
\label{sec:transformer}
We first consider solely inputting the text into our neural network and then discuss how to extend the network to additional numeric/categorical inputs in Section \ref{sec:multimodalnn}. 
While many neural architectures have been proposed to model text, pretrained Transformer networks now dominate modern NLP. These models are first pretrained in an unsupervised manner on a massive text corpus before being fine-tuned over our (smaller) labeled dataset of interest~\citesi{devlin2019bert, raffel2020t5}. 
This allows our supervised learning to benefit from information gleaned from the external text corpus that would otherwise not be available in our limited labeled data. The Transformer also   effectively aggregates information from various aspects of a training example, using a  \emph{self-attention} mechanism to contextualize its intermediate representations based on particularly informative features~\citesi{vaswani2017attention}. 
Since BERT~\citesi{devlin2019bert} first demonstrated the power of Transformer pretraining via Masked Language Modeling (MLM), superior pretraining techniques have been developed. 
\roberta{} \citesi{liu2019roberta} dynamically generates masks and pretrains on a larger corpus for a longer time, employing the same MLM objective as BERT in which random tokens are masked for the  Transformer to guess their original value. 
\electra{}  \citesi{clark2020electra} is an alternative pretraining technique in which a simple generative model randomly replaces tokens and the Transformer must classify which tokens were replaced. 

Given a dataset with multiple text columns, we feed the tokenized text from all columns jointly into our Transformer (with special \textsf{[SEP]} delimiter tokens between fields and a \textsf{[CLS]} prefix token appended at the start \citesi{devlin2019bert}), as detailed in the next paragraph. 
A single embedding vector for all text fields is obtained from the Transformer's representation at the \textsf{[CLS]} position after feeding the merged input into the network~\citesi{devlin2019bert}.   
Similarly, just a single text field can be embedded via the Transformer's vector representation at the \textsf{[CLS]} position, after feeding only this field into the network. 

\begin{figure}[h!]
    \centering
    \includegraphics[width=0.48\textwidth]{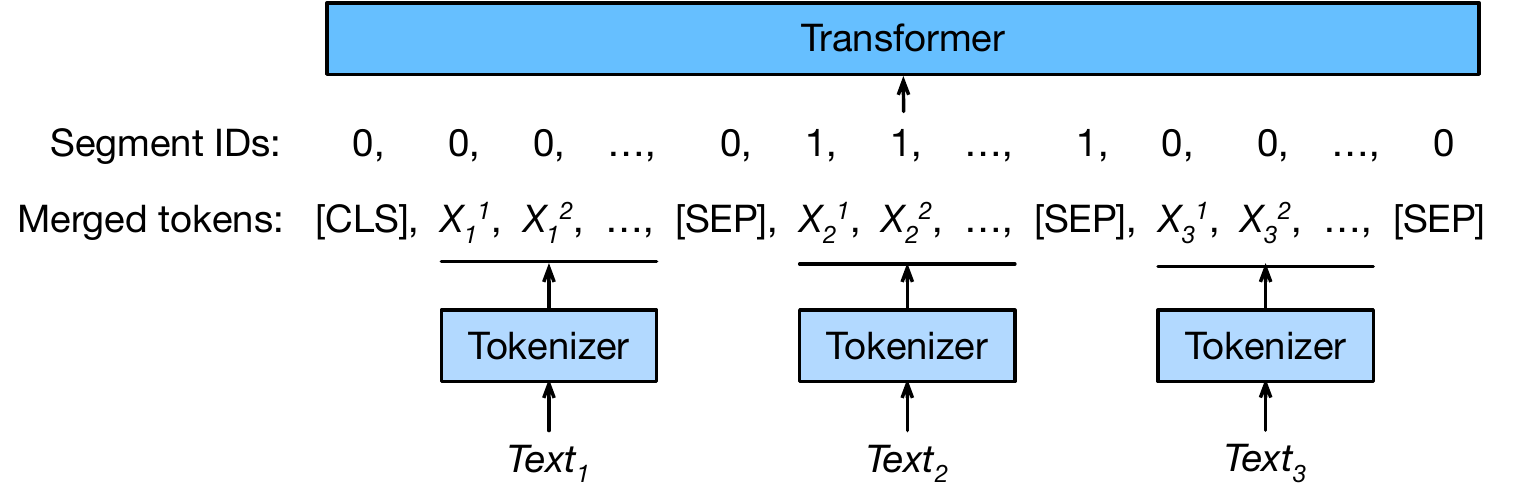} 
    \vspace*{0mm}
     \caption{Inputting data from 3 text fields into Transformer.}
     \label{fig:transformer-input-pipeline}
\vspace*{0em}
\end{figure}

\paragraph{Handling Multiple Text Fields in the Transformer}
Given multiple text columns, we feed the tokenized text from all columns jointly into our Transformer, as illustrated in Figure~\ref{fig:transformer-input-pipeline}. 
We follow the usual method to format text from multiple passages \citesi{devlin2019bert}: tokenized inputs from different text fields are merged with special \textsf{[SEP]} delimiter tokens between fields and a \textsf{[CLS]} prefix token is subsequently appended at the start of merged input. 
To further ensure that the network distinguishes boundaries between adjacent text fields, we alternate $0$s and $1$s as the segment IDs. Here segment IDs and the \textsf{[SEP]} token were previously used to demarcate boundaries between passages during pre-training \citesi{devlin2019bert}. 
After feeding the merged inputs into the Transformer, we can extract its intermediate representations at each position as token-level embeddings (each token has one embedding, which has been contextualized based on information from the other tokens).

When the total length of tokenized text fields exceed the maximum allowed length (set to be $512$ throughout this work), we truncate the input by repeatedly  removing one token from the longest individual text field until the length constraint is met. Since self-attention is permutation equivariant, a common practice is to assign an additional vector that encodes each position (namely positional encoding) so that the Transformer can distinguish between identical tokens occurring at different locations \citepsi{vaswani2017attention}. After merging multiple text fields into a single input, we simply assign positional encodings based on this larger input.

\subsection{Extending Transformer Architectures to Multimodal Inputs}
\label{sec:multimodalnn}

\begin{figure}
    \centering
    \begin{subfigure}[b]{0.45\textwidth}
         \centering
         \includegraphics[width=0.8\textwidth]{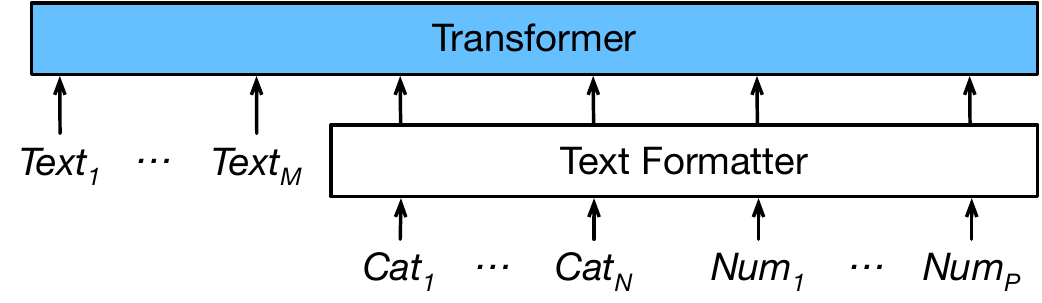}
         \vspace*{-0.3em}
         \caption{\emph{\mmalltext{}}. Convert numeric and categorical values into additional text tokens.}
         \label{fig:fusion-strategy-all-text}
     \end{subfigure}
     \hfill
     \begin{subfigure}[b]{0.45\textwidth}
         \centering
         \includegraphics[width=0.8\textwidth]{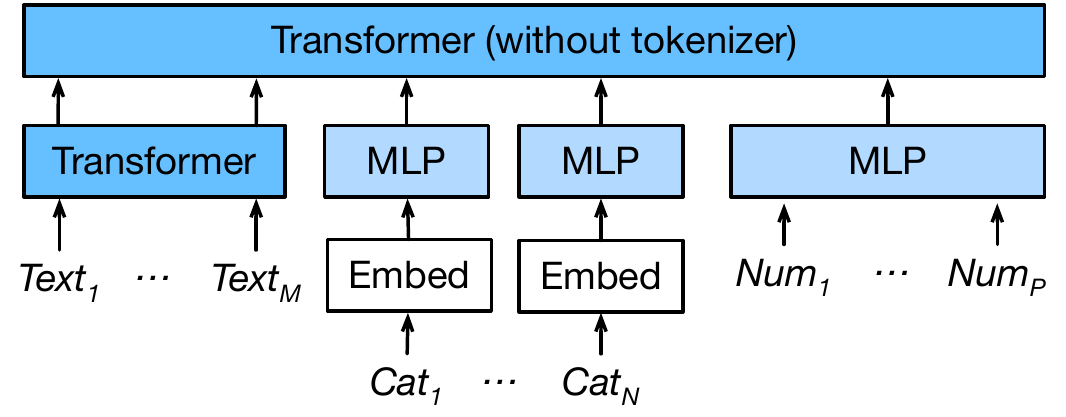}
         \vspace*{-0.3em}
         \caption{\emph{\mmfuseearly{}}. Transformer operates on learned embeddings for each feature. }
         \label{fig:fusion-strategy-early-fusion}
     \end{subfigure}
     \\[0.6em]
    \begin{subfigure}[c]{0.45\textwidth}
         \centering
         \includegraphics[width=0.8\textwidth]{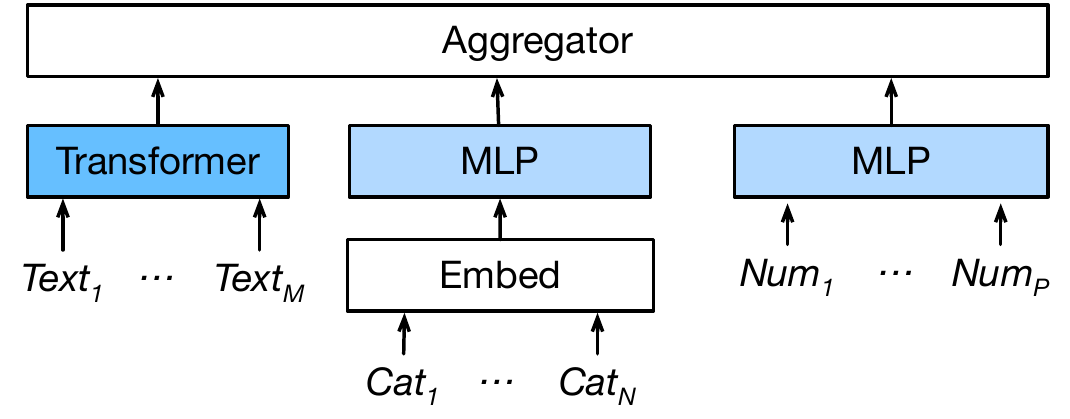}
         \vspace*{-0.3em}
         \caption{\emph{\mmfuselate{}}. Separate branches encode each modality, and aggregated via concatenation. }
         \label{fig:fusion-strategy-late-fusion}
     \end{subfigure}
     \hfill
    \begin{subfigure}[c]{0.45\textwidth}
         \centering
         \includegraphics[width=1.0\textwidth]{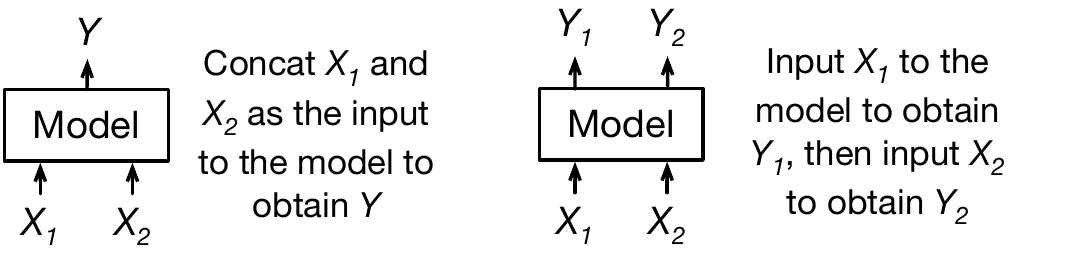}
         \vspace*{-0.3em}
         \caption{Notation used in these figures.}
         \label{fig:fusion-strategy-notation}
     \end{subfigure}
     \vspace*{0em}
    \caption{Options for fusing modalities in \emph{\multimodalnet{}} (Section \ref{sec:multimodalnn}). Two dense layers (not shown) are added on top of each network in (a)-(c) to output a prediction (real value for regression, logit vector for classification). Over our benchmark, option (c) performs best and is the chosen \emph{\multimodalnet{}} architecture that we subsequently try combining with tabular models. 
    }
    \label{fig:fusion-strategy}
\vspace*{-0.5em}
\end{figure}

In many multimodal datasets, some of the predictive signal solely resides in text fields, while other predictive information is restricted to tabular feature values, or complex interactions between text and tabular values. 
To enjoy the benefits of end-to-end learning without sacrificing accuracy, we consider how to adapt a Transformer network to simultaneously operate on inputs from both modalities, referring to the resulting network as \emph{\multimodalnet}. 
A natural approach in our setting is to enhance the Transformer such that its attention mechanism can contextualize representations of individual text tokens based not only on other parts of the text, but also on the values of relevant tabular features as well.  Below we discuss three different options for implementing the  \emph{\multimodalnet{}} that are depicted in Figure \ref{fig:fusion-strategy} (with details in Appendix~\ref{sec:appdx-network_architecture}). These options differ in whether  information is fused across text and tabular modalities: at the input layer (\emph{\mmalltext{}}), in the earlier layers of the network near the input (\emph{\mmfuseearly{}}), or in the later layers of the network near the output (\emph{\mmfuselate{}}).

\vspace*{\preparagraphspaces{}}
\paragraph{\mmalltext{}} A simple (yet crude) option is to convert numeric and  categorical values to strings and subsequently treat their columns also as text fields \citesi{raffel2020t5}. 
Through its byte-pair encoding, a pretrained Transformer can handle most categorical strings and may be able to crudely represent numeric values within a certain range (here we round all numbers to 3 significant digits in their string representation).

\vspace*{\preparagraphspaces{}}
\paragraph{\mmfuseearly{}} Rather than casting them as strings, we can allow our model to adaptively learn token representations for each numeric and categorical feature via backpropagation (see Figure~\ref{fig:fusion-strategy-early-fusion}). 
We introduce an extra factorized embedding layer~\citesi{lan2019albert,guo2016entity} to map categorical values into the same $\mathbb{R}^d$ vector representation encoded by the pretrained Transformer backbone for text tokens (with different embedding layers used for different categorical columns in the table). 
All numeric features are encoded via a single-hidden-layer Multi-layer Perceptron (MLP) to obtain a unified $\mathbb{R}^d$ vector representation. The resulting $d$-dimensional vector representations from each modality are jointly fed into a 6-layer Transformer encoder whose self-attention operations can model interactions between the embeddings of text tokens, categorical values, and numeric values. 
We refer to this strategy as \emph{Fuse}-\emph{Early} because only a minimal (yet adaptive) input processing layer is added to convert the tabular features into a common vector form which can be jointly fed through many shared Transformer layers. 
\citetsi{huang2020tabtransformer} considered a similar strategy for applying Transformers to entirely numeric/categorical data, albeit without text components that are a major focus here.

\vspace*{\preparagraphspaces{}}
\paragraph{\mmfuselate{}} 

Rather than aggregating information across modalities in early network layers, we can perform separate neural operations on each data type and only aggregate per-modality representations into a single  representation near the output layer (see Figure \ref{fig:fusion-strategy-late-fusion}). 
This multi-branch design allows each branch to extract higher-level representations of the values from each modality, before the network needs to consider how modalities should be fused. 
Here we use a multi-tower architecture in which numeric and categorical features are fed into separate MLPs for each modality. The text features are fed into a (pretrained) Transformer network. The topmost vector representations of all three networks are pooled into a single vector from which predictions are output via two dense layers. 
As pooling operators, we considered mean/max pooling or concatenation as options. Experiments show these pooling methods perform similarly on each dataset, with concatenation exhibiting  slightly better overall performance, and we thus fix concatenation as our pooling method in the \mmfuselate{} architecture.

\subsection{Featurizing Text for Tabular Models}
\label{sec:featurizetext}

Despite their success for modeling text, the application of Transformer architectures to tabular data remains limited \citesi{huang2020tabtransformer,fakoor2020trade, fakoor2020fast}. 
The use of tabular models together with Transformer-like text architectures has also received little attention \citesi{wan2020nbdt, ke2019deepgbm}.  
Recall that `tabular models' throughout are those trained on only numeric/categorical features, e.g.\  different types of decision tree ensembles fit by AutoGluon-Tabular. 

To allow tabular models to access information in text fields, the text is typically first mapped to a continuous vector representation which replaces a text column in our data table with multiple numeric columns (one for each vector dimension).  One can treat each text column as a document, and each individual text field as a paragraph within the document, such that each text field can be featurized via NLP methods for computing text representations \citesi{eisenstein2018natural, mikolov2013efficient, shan2016deep} before the tabular models are trained. 

Rather than classical NLP methods like N-grams or word embeddings \citesi{eisenstein2018natural}, a Transformer can instead be used to map the text fields into a vector representation via contextual embedding \citesi{devlin2019bert, blohm2020leveraging}. Subsequently, the text fields are replaced in the data table by additional numeric columns corresponding to each dimension of the embedding vector (Figure~\ref{fig:embed-as-feature}). Our study considers three ways to featurize text using a Transformer.

\paragraph{\preembed{}}  Most straightforward is to embed text via a pretrained Transformer (not fine-tuned on our labeled data), and  subsequently train tabular models over the featurized data table  \citesi{blohm2020leveraging}.

\vspace*{\preparagraphspaces{}}
\paragraph{\textembed{}} The \emph{\preembed{}} strategy is not informed about our particular prediction problem and the domain of the text data. In \emph{\textembed{}}, we further fine-tune the pretrained Transformer to predict our labels from only the text fields, and use the resulting Text-Net to embed the text. By adapting to the domain of the specific prediction task, \emph{\textembed{}} is able to extract more relevant textual features that can improve the performance of tabular models. 
This is particularly true in settings  where the target only depends on one out of many text fields, since the fine-tuning process can produce representations that  vary more based on the relevant field vs.\ irrelevant text.

\vspace*{\preparagraphspaces{}}
\paragraph{\multimodalembed{}} Text representations may improve when  self-attention is informed by context regarding numeric/categorical features. Thus we also consider embedding text via our best  multimodal network from Section \ref{sec:multimodalnn} (depicted in Figure \ref{fig:fusion-strategy-late-fusion}). These models are again fine-tuned using the labeled data and now produce a single vector representation for \emph{all} columns in the dataset, regardless of their type. Since Transformers are better suited for modeling text than tabular features, we only replace the text fields with the learned vector, all other non-text features are kept and used for subsequent tabular learning. Thus the sole difference between \emph{\textembed{}} and \emph{\multimodalembed{}} is that the embeddings used to replace text are additionally contextualized on numeric/categorical feature values in the latter method.

\subsection{Aggregating Text \& Tabular Models}
\label{sec:ensembling-methods}

\begin{figure}[tb!]
    \centering
    \begin{subfigure}[b]{0.32\textwidth}
         \centering
         \includegraphics[width=0.8\textwidth]{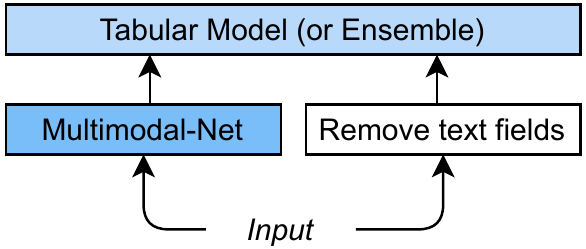}
         \caption{\emph{Embedding-as-Feature}}
         \label{fig:embed-as-feature}
     \end{subfigure}
     \hfill
    \begin{subfigure}[b]{0.32\textwidth}
         \centering
         \includegraphics[width=0.8\textwidth]{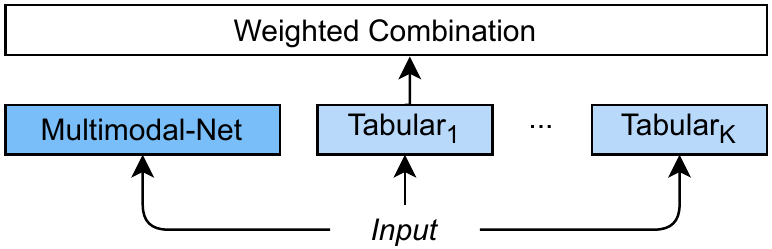}
         \caption{\emph{\mmlinearensemble{}} }
         \label{fig:linear-ensemble}
     \end{subfigure}
     \hfill
     \begin{subfigure}[b]{0.32\textwidth}
         \centering
         \includegraphics[width=0.8\textwidth]{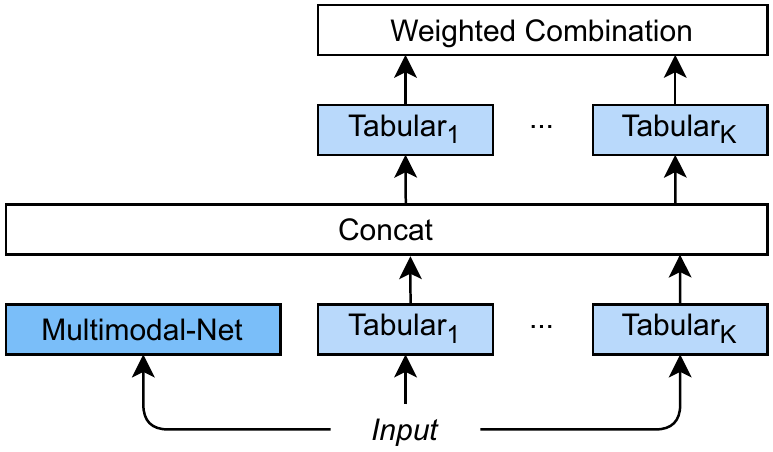}
         \caption{\emph{\mmstackensemble{}} }
         \label{fig:stack-ensemble}
     \end{subfigure}
     \vspace*{0em}
     \caption{Options for combining \emph{\multimodalnet{}} with classical tabular models. Five particular tabular models are used in this paper: extremely randomized trees, a simple  MLP, and three different types of gradient boosted decision trees.  
     Over our benchmark, option (c) performs the best and is chosen as the strategy for aggregating text and tabular models in our proposed AutoML solution. 
     }
     \label{fig:aggregation}
     \vspace*{-0em}
\end{figure}

Rather than merely leveraging the Transformers for their embedding vector representations as in Section \ref{sec:featurizetext}, an alternative multimodal text/tabular modeling strategy is to instead consider their predictions and ensemble these with predictions from tabular models. 
Utilized by most AutoML frameworks \citesi{ledell2020h2o,feurer2015efficient,agtabular}, model ensembling is a straightforward technique to boost predictive accuracy. 
Ensembling is particularly suited for multimodal data, where different models may be trained with different modalities. However, the resulting ensemble may then be unable to exploit nonlinear predictive interactions between features from different modalities. To remedy this, we advocate for the use of our multimodal Transformers (from Section \ref{sec:multimodalnn}) that fuse information from  text and tabular inputs. Here we specifically consider ensembling the multimodal Transformer model with the various standard tabular models used by AutoGluon-Tabular. 
Furthermore, we propose stack ensembling with nonlinear aggregation of model predictions that can exploit inter-modality interactions between different base models' predictions, even when  base models do not overlap in modality.

\vspace*{\preparagraphspaces{}}
\paragraph{\mmlinearensemble{}} 

We first consider straightforward aggregation via a weighted average of the predictions from our Transformer model and various tabular models (like those trained by AutoGluon-Tabular). 
Here, our Transformer and other models are independently trained using a common training/validation split. 
Subsequently, we apply \emph{ensemble selection}, a forward-selection algorithm to fit aggregation weights over all models' predictions on the held-out validation data \citesi{caruana2004ensemble}. 
Unlike regression for fitting the aggregation weights \citesi{van2007super, ledell2020h2o}, ensemble selection is favored by many tabular AutoML tools like AutoGluon as it is more computationally efficient, less prone to overfitting, and naturally favors sparse weights 
\citesi{feurer2019auto, agtabular}. 

\vspace*{\preparagraphspaces{}}
\paragraph{\mmstackensemble{}} 
Rather than restricting the aggregation to a linear combination, we can use stacking   \citesi{wolpert1992stacked}. This trains another ML model to learn the best aggregation strategy. The features upon which the  `stacker' model operates are the predictions output by all base models (including our Transformer), concatenated with the original tabular features in the data. 
Following \citetsi{agtabular}, we try each type of tabular model in AutoGluon-Tabular as a stacker model (see Appendix \ref{sec:ag-model-stacker}).  To output predictions, a weighted ensemble is constructed via ensemble selection applied to the tabular stacker models (Figure~\ref{fig:stack-ensemble}). We do not consider our larger (multimodal) Transformer model as a stacker since lightweight aggregation models are preferred in practice. 
Overfitting is a key peril in stacking, and we ensure that stacker models are only trained over \emph{held out} predictions produced from base models via $5$-fold cross-validation (bagging) \citepsi{van2007super, agtabular}.

\section{Additional Experiment/Implementation Details}
\label{sec:model_details}

\begin{table*}[h!]
\begin{center}
{\tiny
    \centering
    \begin{tabular}{lccccccc}
        \toprule
        Method & binary & multiclass & regression \\
        \midrule
\roberta             &   0.843 &       0.511 &       0.501 \\

\electra            &   0.823 &       0.545 &       0.519 \\
+ Exponential Decay $\tau=0.8$ &   0.885 &       0.545 &       0.544 \\
        + Average 3 $\bigstar$  &   0.887 &       0.548 &       0.546 \\
        \midrule
         \gray{Choosing \multimodalnet{}:} &  \multicolumn{3}{c}{Fusion Strategy } \\
\mmalltext{} &   0.893 &       0.641 &       0.548 \\
\mmfuseearly{}  &   0.885 &       0.643 &       0.546 \\
\mmfuselate{} $\bigstar$ &   0.889 &       0.649 &       0.551 \\
        \midrule
        \gray{Choosing Aggregation:} & \multicolumn{3}{c}{Multimodal Model Aggregation } & & & \\
\preembed{} &   0.781 &       0.638 &       0.430 \\
 \textembed{} &   0.798 &       0.657 &       0.528 \\
        \multimodalembed{} &   0.799 &       0.673 &       0.532 \\
        \mmlinearensemble{} &   0.886 &       0.682 &       0.549 \\
        \mmstackensemble{} $\bigstar$ &   \textbf{0.902} &       \textbf{0.690} &       \textbf{0.553} \\
        \midrule
        \gray{Baselines:} & \multicolumn{3}{c}{Tabular AutoML + Feature Engineering } & & \\
        \agweighted &   0.697 &       0.510 &       0.154 \\
        \agstack &   0.700 &       0.513 &       0.155 \\
        \agweighted + \ngram  &   0.872 &       0.679 &       0.471 \\
        \agstack + \ngram &   0.877 &       0.688 &       0.519 \\
        \hto &   0.692 &       0.551 &       0.343 \\
        \htow2v &   0.843 &       0.627 &       0.445 \\
        \htoemb &   0.769 &       0.567 &       0.427 \\
        \bottomrule
    \end{tabular}
    }
    \caption{Alternative summary of AutoML results over our multimodal benchmark, where performance on each dataset is separately averaged over the \textbf{binary} classification tasks (i.e.\ average AUC), \textbf{multiclass} classification tasks (i.e.\ average accuracy), and \textbf{regression}  tasks (i.e.\ average $R^2$). See Table \ref{tab:main} for additional details. }
    \end{center}
    \end{table*}

\subsection{Data Processing}
For tabular features/models, we can simply rely on the same  preprocessing as AutoGluon-Tabular, which has been found to also work well for other learning algorithms \citepsi{agtabular}. 
For our subsequently introduced multimodal neural networks that operate on both text and tabular features, we simply rescale and center numeric features and impute their missing values via their average. Missing values for categorical features (and previously unseen categories encountered during inference) are represented via an additional \textsf{Unknown} category in order to avoid unrealistic missing at random assumptions. Missing text fields are handled as empty strings in our preprocessing pipeline.  
The tabular MLP networks in AutoGluon-Tabular (the AutoML solution around which our experiments are based) also only use this simple preprocessing. We thus also utilized the same preprocessing for other neural networks evaluated in our experiments for controlled comparison. Note this was only done for the experiments presented here; the actual datasets in our benchmark have not been preprocessed in this manner, and the benchmark leaves preprocessing as a challenge for future AutoML systems to address as they see fit.


AutoGluon also automatically infers the type of each feature  via simple yet effective heuristics. 
One decision particular to our multimodal applications is when to designate a column of string values as a categorical vs.\ text feature. In this work, we simply threshold based on the number of unique values in the column, such that commonly reoccurring strings are treated as discrete categories rather than unstructured text. We choose the threshold to be 20 in all presented experiments, based on visually confirming the inferred feature types with this threshold agree with our intuition regarding which columns should be handled as text.

While the aforementioned steps are used to report the feature types for each dataset listed in Table \ref{tab:benchmark},  we note that our benchmark does \emph{not} require systems to treat certain columns as particular data types. Feature type inference is instead left up to individual methods, since  automatically identifying the best way to treat certain columns remains an important research question.  


\subsection{Network Architectures}
\label{sec:appdx-network_architecture}

In this paper, we used a single-hidden-layer MLP as the basic building block for encoding features and projecting the hidden states.
It has one bottleneck layer and uses layer normalization. We use the leaky ReLU activation (with slope set to $0.1$) for all basic MLP layers mentioned throughout  the paper. For the 6-layer Transformer model in \emph{\mmfuseearly{}}, we used the GeLU activation like \citet{devlin2019bert}. We set the number of units, heads, and hidden size of FFN (the feedforward layers) in this Transformer to be 64, 4, 256 correspondingly. For the categorical features, we use an encoding network that is similar to the factorized embedding in ALBERT~\citepsi{lan2019albert}, in which we use an embedding layer with $32$ units and then project it with a basic MLP layer that has 64 bottleneck units. We further set the number of output units in the basic MLP to be the same as the token-embeddings used in the pretrained Transformer model (i.e., \electra{}  or \roberta) so that all vectors belong to the same space. 

In the \emph{\mmfuselate{}} variant, we further concatenate all encoded categorical features and encode them with a second basic MLP layer. Numeric features are concatenated and encoded with one basic MLP layer. These MLP layers all utilize 128  bottleneck units and their output unit number matches the dimensionality of token embeddings for the pretrained Transformer. 
The total number of parameters for the \emph{All-Text}, \emph{Fuse-Late}, and \emph{Fuse-Early} multimodal network variants are: 109.0 million, 109.1 million, and 109.3 million correspondingly. Thus, these three model variants have comparable costs.

\subsection{Neural Network Optimization}
\label{sec:appdx-optimization}
All text/multimodal neural networks are trained with the slanted triangular learning rate scheduler~\citepsi{howard2018universal} with initial learning rate set to 0.0, the maximal learning rate set to $5 \times 10^{-5}$ and warmup set to $0.1$. We use a  batch size of $128$, $10^{-4}$ weight decay, and the AdamW optimizer. Text/multimodal networks are trained for 10 epochs and we early stop based on their validation performance. 
These learning rate and weight decay values were determined via grid search on a single smaller (subsampled) dataset that we used for early initial experiments. 


\subsection{Details of AutoGluon Tabular Models in the Stack Ensemble}
\label{sec:ag-model-stacker}
For better efficiency, we considered just the following tabular models when running AutoGluon~\citepsi{agtabular}:
\begin{itemize}
\item Fully-connected Neural Network (MLP) with ReLU activations~\citepsi{agtabular}.

\item LightGBM model with default hyperparameters (GBM)~\citepsi{ke2017lightgbm}.
\item A second LightGBM model with a different set of hyperparameter values. By default, AutoGluon  uses this second model in conjunction with the first LightGBM model. 
\item An implementation of Extremely Randomized Trees from the LightGBM library \citepsi{geurts2006extremely}.
\item CatBoost gradient boosted trees for sophisticated handling of categorical features~\citepsi{dorogush2018catboost}.
\end{itemize}
To avoid overfitting in stacking, all models are trained with 5 fold cross-validation (bagging) as described by \citet{agtabular}. For classification tasks, the outputs of each base model which are aggregated in the ensemble are taken to be predicted class probabilities.  

\subsection{Notes on Hyperparameter Tuning}
\label{sec:hpo}

Note that hyperparameter tuning was not a major focus in the preliminary study conducted in this paper. Standard hyperparameter tuning strategies \citepsi{shahriari2015taking} are readily applicable to our multimodal setting, and the experiments presented here could easily employ the advanced Bayesian optimization techniques available in AutoGluon \citepsi{tiao2020model}. 
We expect the performance of all of our proposed AutoML strategies will grow even better with time devoted to hyperparameter tuning. 
However in this paper we did not conduct such a search and simply used the default hyperparameters supplied by AutoGluon for tabular models, which are already highly performant \citepsi{agtabular}, and the text/multimodal network hyperparameters are listed here and are viewable in our released code. Over just a few datasets, we found that relative performance of different strategies did not qualitatively differ with other reasonable manually-chosen hyperparameter settings (i.e.\ hyperparameter values known to generally work well for these specific models such as alternative popular learning rate schedules or small changes to the size of the networks).

Rather than only reporting a couple thoroughly-tuned results, we instead preferred to spend our time/compute budget to explore more modeling strategies over more datasets. 
Note that all H2O AutoML variants reported in Table \ref{tab:main} relied on extensive hyperparameter sweeps (automatically used within H2O), and yet were still unable to outperform some of the other untuned methods we considered. This further supports the claim that our benchmark has helped us identify a broadly performant strategy for multimodal AutoML. 

\subsection{Compute Details}
\label{sec:compute_details}

All experiments were run on Amazon Web Services EC2 cloud instances (P3.2xlarge). Each instance has two NVIDIA V100 Tensor Core GPUs. About 2000 hours of total compute was required for all experiments presented in this paper (18 instances used for about a week). 
Given a limited compute budget, we believe more meaningful conclusions may be drawn by  running more algorithms over more datasets rather than replicate runs of different seeds/splits on just a few (less diverse) datasets. We also did not include any small datasets in our benchmark for which replicate runs would otherwise be required to get statistically stable results.

\clearpage 
\section{Dataset Details}
\label{sec:datasetdescriptions}

Each dataset can be easily loaded into Python (or another programming language) as the  standard dataframe format used by \texttt{pandas}. All tables in our benchmark are appropriately formatted for supervised learning, with the first row serving as a standard header whose columns specify the names of each feature. 
We release our modified versions of the datasets in our benchmark under a \textbf{CC BY-NC-SA} license, and note that any data from this benchmark which has previously been published elsewhere falls under the original license from which the data originated (links to the original sources are provided). We the authors bear all responsibility in case of violation of rights. Long-term preservation of our benchmark is ensured by hosting the repository on GitHub, such that users can contribute their own improvements or publicly raise issues for us to address. The data files are hosted in AWS Simple Cloud Storage (S3), a reliable medium that will ensure researchers can easily obtain these files. 
Appendix \ref{sec:datasheet} provides a datasheet \citesi{gebru2018datasheets} for our overall benchmark.

\noindent
\textbf{prod}: Classify the sentiment (4-way classification) of user reviews of products based on the review text and product type (e.g.\ Tablet, Mobile, etc.). Intuitively, we expect most of the predictive signal to lie in the text, but predictions can be further improved by accounting for the fact that certain types of products tend to receive certain user sentiment.  
Representing a relatively simple multimodal task with only a single text feature and one categorical feature, this dataset originally stems from a 2020 MachineHack prediction  competition: 
\url{https://machinehack.com/hackathons/product_sentiment_classification_weekend_hackathon_19/overview}
\\[\itemspaces{}]

\noindent
\textbf{salary}: Predict the salary range 
listed in data scientist job postings (in India) given the job description as well as other features like skill requirements and location. Intuitively, the best models will learn to identify valuable requirements from the text and high salary locations (via categorical modeling) as well as predictive interaction-effects. 
Representing a task with many text fields, this dataset originally stems from a 2018 MachineHack prediction competition:
\url{https://machinehack.com/hackathons/predict_the_data_scientists_salary_in_india_hackathon/overview}
\\[\itemspaces{}]

\noindent
\textbf{airbnb}: Predict the price label of AirBnb listings (in Melbourne, Australia) based on information from the listing page including various text descriptions and many numeric features (e.g.\ host's response-rate, number of bed/bath-rooms) and categorical features (e.g.\ property type, superhost or not). 
Representing a complex classification task with many features from each modality, the original version of this dataset was released via the InsideAirbnb initiative:  
\url{https://www.kaggle.com/tylerx/melbourne-airbnb-open-data} 
\\[\itemspaces{}]

\noindent
\textbf{channel}: Predict which news category (i.e.\ channel) a Mashable.com news article belongs to based on the text of its title, as well as auxiliary numerical features like the number of words in the article, its average token length, how many keywords are listed, etc. 
Representing a task with one text field but many tabular (numeric) features, the original version of this dataset was collected by \citesi{fernandes2015proactive}: 
\url{https://archive.ics.uci.edu/ml/datasets/online+news+popularity}
\\[\itemspaces{}]

\noindent
\textbf{wine}: Classify the variety of wines based on tasting descriptions from sommeliers, and numeric features like price and categorical features like country-of-origin. The original version of this dataset was collected from WineEnthusiast: 
\url{https://www.kaggle.com/zynicide/wine-reviews}
\\[\itemspaces{}]

\noindent
\textbf{imdb}: Predict whether or not a movie falls within the Drama category based on text features like its name, description, actors/directors, and numerical features like its release year, runtime, etc. 
Representing a task with smaller sample-size, the original version of this dataset was collected from IMDB (the most popular movies): 
\url{https://www.kaggle.com/PromptCloudHQ/imdb-data}\symfootnote{\label{note:prompt}PromptCloud released the original version of the data from which we created this benchmark dataset.} 
\\[\itemspaces{}]

\noindent
\textbf{fake}: Predict whether online job postings are real or fake based on their text  and additional tabular features like amount of salary offered and degree of education required. 
Representing an imbalanced binary classification task, these data stem from the Employment Scam Aegean Dataset collected by \citesi{vidros2017automatic}: 
\url{https://www.kaggle.com/shivamb/real-or-fake-fake-jobposting-prediction}
\\[\itemspaces{}]

\noindent
\textbf{kick}: Predict whether a proposed Kickstarter project will achieve funding goal based on text features like its title, description, numeric features like the amount of money requested, date posted, and categorical features like the country, currency, etc. 
This dataset represents a complex task where models must consider interactions between modalities to address a core question of Kickstarter's business:  
\url{https://www.kaggle.com/codename007/funding-successful-projects}
\\[\itemspaces{}]

\noindent
\textbf{jigsaw}: Predict whether online social media comments are toxic based on their text and additional tabular features providing information about the post (e.g.\ likes, rating, date created, etc.). 
This dataset originates from a 2019 Kaggle competition (\url{https://www.kaggle.com/c/jigsaw-unintended-bias-in-toxicity-classification}) in which the 1st place solution\footnote{\scriptsize  \url{https://www.kaggle.com/c/jigsaw-unintended-bias-in-toxicity-classification/discussion/103280}} utilized dataset-specific tricks such as a Bucket Sequencing Collator, auxiliary domain-specific prediction tasks for models, and a custom mimic loss function for training. 
\\[\itemspaces{}]

\noindent
\textbf{qaa}: Given a question and an answer (from the Crowdsource team at Google) as well as an additional category feature, predict the (subjective) type of the answer in relation to the question. 
Representing a predominantly NLP task that requires deep language understanding (though the most accurate models must also consider the category), this dataset stems from a 2019 Kaggle competition:  
\url{https://www.kaggle.com/c/google-quest-challenge}
\\[\itemspaces{}]

\noindent
\textbf{qaq}: Given a question and an answer (from the Crowdsource team at Google) as well as additional category features, predict the (subjective) type of the question in relation to the answer. These data stem from the same source as \textbf{qaa}, where the different labels were both prediction targets in the original (multi-label) Kaggle competition.   
 \\[\itemspaces{}]
 
 \noindent
\textbf{book}: Predict the sale price of books based on text features like their title, author, synopsis, categorical features like genre and numeric features like customer reviews and overall rating. This dataset originally stems from a 2019 MachineHack prediction competition: 
\url{https://machinehack.com/hackathons/predict_the_price_of_books/overview}
\\[\itemspaces{}]

\noindent
\textbf{jc}: Predict the sale price of items sold on the website of the retailer JC Penney based on text features like its title/description, and numeric features like its rating.  
Representing an important (e)commerce task, this data was originally collected using information from the online page for each product:  
\url{https://www.kaggle.com/PromptCloudHQ/all-jc-penny-products}*
\\[\itemspaces{}]

\noindent
\textbf{cloth}: Predict the score of a customer review of clothing items (sold by an anonymous retailer) based on the review text, how much positive feedback the review has received (numeric), and additional features about the product like its department (categorical). 
The data were collected by \citesi{agarap2018statistical}: 
\url{https://www.kaggle.com/nicapotato/womens-ecommerce-clothing-reviews}
\\[\itemspaces{}]

\noindent
\textbf{ae}: Predict the price of inner-wear items sold by retailer American Eagle based on  text features like their product name, description, categorical features like brand, and numeric features like rating, review count. 
Representing an important (e)commerce task, this data was originally collected using information from the online page for each product: 
\url{https://www.kaggle.com/PromptCloudHQ/innerwear-data-from-victorias-secret-and-others}*
\\[\itemspaces{}] 

\noindent
\textbf{pop}: Predict the popularity (number of shares on social media, on log-scale) of Mashable.com news articles based on the text of their title, as well as auxiliary numerical features like the number of words in the article, its average token length, and how many keywords are listed, etc. 
This dataset represents a very difficult prediction problem with only weak signal offered by the observed features. It is fundamentally hard to forecast how popular an article will be based only on its title and crude numerical summary statistics. To be comprehensive, an AutoML benchmark should contain at least one challenging problem like this. 
While \textbf{pop} stems from the same original data source as \textbf{channel}, the two have different labels to predict and do not share exactly the same set of features. 
\\[\itemspaces{}]

 \noindent
\textbf{house}: Predict sale prices of California homes sold in 2020 based on a text summary written by the seller and various tabular features 
(e.g.\ bedroom number, home type, location, year built, parking). 
Representing a regression task with many features that are text and numeric, this dataset originally stems from a 2021 Kaggle prediction competition: 
\url{https://www.kaggle.com/c/california-house-prices} 
\\[\itemspaces{}]

\noindent
\textbf{mercari}: Predict the price of items sold in the online marketplace of Mercari based on information from the product page like name, description, free shipping availability, etc. 
This data originates from a 2017 Kaggle competition  (\url{https://www.kaggle.com/c/mercari-price-suggestion-challenge/}), in which 1st  place\footnote{\scriptsize   \url{https://www.kaggle.com/c/mercari-price-suggestion-challenge/discussion/50256}} and 3rd place\footnote{\scriptsize  \url{https://www.kaggle.com/c/mercari-price-suggestion-challenge/discussion/50272}}  engineered dataset-specific text features such as customized bag-of-words  
and character N-grams,  
carefully tuned learning-rate/batch-size schedules, and specially ensembled models in a dataset-specific manner. 
\\[\itemspaces{}]

\subsection{Datasheet for our Multimodal Text/Tabular Benchmark}
\label{sec:datasheet}

To avoid redundancy, we only provide details here not covered elsewhere in the paper or our benchmark repository. Table \ref{tab:benchmark} lists statistics of each dataset. For details on how each dataset was collected, please refer to the original source linked in our benchmark repository. 
\\[\itemspaces{}]

\textbf{How were datasets selected for the benchmark?}
The 18 datasets in our benchmark represent all of the public text/tabular datasets we could find that do not violate our exclusion criteria and satisfy our main desiderata: the dataset must entail a meaningful prediction problem with real enterprise data (as opposed to contrived toy task without real-world application). Note that we only consider tabular datasets that contain text fields, which is a small fraction of publicly available tabular datasets (even though such data are ubiquitous in private enterprises). Our dataset search was conducted over the following sources: Kaggle, MachineHack, UCI ML Repository; the first two are the best sources of publicly available enterprise datasets (with meaningful prediction problems) that we are aware of.

Within each source, we searched for datasets matching the keyword ``text'' in their metadata/descriptions for consideration in our benchmark (although the majority such datasets either had no tabular features or failed to provide the original raw text presenting only a featurized version such as bag-of-words). We also conducted some dataset searches via Google, but did not find serious candidates for the benchmark via this avenue. Beyond the primary requirement that data must stem from a real enterprise application with a meaningful classification/regression task, our other exclusion criteria ensured each dataset in the benchmark has: IID examples, non-prohibitive licensing, some text fields beyond just 1-2 words and in the English language (for simplicity), sample size of at least 1000, and predictive signal across both text and tabular (numeric+categorical) modalities (meaning one modality does not appear entirely useless for the prediction problem, evaluated via preliminary \emph{AG-Stack+Ngram} runs without each modality).

\paragraph{For what purpose were the benchmark datasets created?} We collected the datasets in this benchmark to evaluate supervised machine learning (classification/regression) algorithms designed to jointly operate on text and tabular features. The original versions of these data were also initially created primarily for a similar purpose. 
\\[\itemspaces{}]

\paragraph{Who created this benchmark? Who funded its creation?}
The authors of this paper, all scientists employed by Amazon, curated this benchmark. Curating the benchmark did not cost significant money, and the benchmark data are currently hosted on cloud servers (S3) provided by Amazon. 
The original data sources were created/curated/funded by various companies/individuals, please refer to each individual source for more details. 
\\[\itemspaces{}]

\paragraph{Do the datasets contain all possible instances or are they a sample (not
necessarily random) of instances from a larger set?}
Each dataset is a sample of instances from a larger set. We caution these samples may not be at all representative of the larger set, and thus the benchmark should not be used to draw  domain-specific conclusions/insights through scientific data analysis of individual datasets.
\\[\itemspaces{}]

\paragraph{Is any information missing from individual instances?}
Yes there are many missing fields in certain datasets. It is unclear why they are missing or if the missingness mechanism satisfies the missing at random assumption.
\\[\itemspaces{}]

\paragraph{Are relationships between individual instances made explicit?} 
For evaluating ML performance, we simply assume the data are IID. However this may be violated by certain datasets. For example, product datasets may contain near duplicate products and products may be related (reviewed by the same users, price of a product can affect price of others, etc.). We do not explicitly know the relationships between instances in these data.
\\[\itemspaces{}]

\paragraph{Are there recommended data splits (e.g., training, development/validation, testing)?} 
Yes the benchmark provides a recommended training/test split, but ML systems are free to split validation data from the training set as they see fit. The split was done randomly (stratified based on labels for classification) to best reflect an IID setting for which supervised learning methods are primarily intended.  
\\[\itemspaces{}]

\paragraph{Does the benchmark contain data that might be considered confidential?} 
Not to our knowledge, but it is possible that a person entered confidential information into the text fields (although they knew these would be publicized).
\\[\itemspaces{}]

\paragraph{Does the benchmark contain data that, if viewed directly, might be offensive, insulting, threatening, or might otherwise cause anxiety?}
The data are mostly non-offensive data used for business purposes. Exceptions are the text fields in the \emph{jigsaw} dataset, which contain toxic online comments, and the \emph{channel}/\emph{pop} datasets, which contain news article titles that may be anxiety-inducing. Furthermore, some of the user reviews of products may be offensive to certain people, although we did not spot any. 
\\[\itemspaces{}]

\paragraph{Does the benchmark relate to people?}
Yes some datasets contain information from people. These all stem from commercial sources where people upload their data intentionally to share it with the world (e.g.\ user reviews, Kickstarter fundraising, public questions, etc.). There is no sensitive/personal information in these data, beyond what a person intended to publicize. 
\\[\itemspaces{}]

\paragraph{Is it possible to identify individuals, either directly or indirectly from the benchmark?}
Yes it may be possible as some datasets contain text fields where an individual may have entered arbitrary information (although they knew the information would appear publicly). 
\\[\itemspaces{}]

\paragraph{Does the benchmark contain data that might be considered sensitive in
any way?} 
Not to our knowledge given all this data was already publicly available, but it is possible given the nature of free form text fields.
\\[\itemspaces{}]

\paragraph{How did you process the data from the original sources? Is the software used to preprocess/clean the datasets available?} 
We processed each dataset from the original source using the publicly available scripts in the \textbf{scripts/data\_processing/} folder of our benchmark GitHub repository. To create versions for our benchmark, we omitted certain features (columns), badly formatted or duplicated rows and subsampled overly large datasets. 
\\[\itemspaces{}]

\paragraph{Have the benchmark data been used for any tasks already?}
Yes many of the datasets have been used to evaluate ML systems, some through formal prediction competitions. Other datasets have been used to demonstrate data analysis techniques. For the datasets originally stemming from Kaggle, one can find some of the previously considered tasks in the discussion forum or notebooks associated with the original dataset. 
\\[\itemspaces{}]

\paragraph{What (other) tasks could the benchmark data be used for? Are there tasks for which these data should not be used?}
We recommend these datasets only be used for evaluation of machine learning algorithms. One could select different target variables in each dataset to create new prediction tasks to evaluate, but these will likely be less practically meaningful (i.e.\ representative of a real application) than the target variable we have selected for each dataset. Also note that none of the datasets has extremely large sample-size (say over a million), so modeling conclusions drawn based on this benchmark may not translate to applications with massive datasets.
\\[\itemspaces{}]

\paragraph{Will the benchmark be distributed to third parties outside of the entity on behalf of which the dataset was created?} 
Yes the benchmark is made publicly available. 
\\[\itemspaces{}]

\paragraph{Have any third parties imposed IP-based or other restrictions on the
data?} 
Yes please refer to the licenses corresponding to each original data source (linked from our repository) for more details.
\\[\itemspaces{}]

\paragraph{Do any export controls or other regulatory restrictions apply to the
dataset or to individual instances?} 
Not to our knowledge.
\\[\itemspaces{}]

\paragraph{How can the curators of the benchmark be contacted?}
You can open a GitHub issue at the benchmark repository, or email the authors of this paper.

\paragraph{Will the benchmark be updated (e.g.\ to correct errors, add
new datasets, add/delete instances)?} 
Yes updates will be done via GitHub and publicly announced there.
\\[\itemspaces{}]

\paragraph{If others want to extend/augment/build on/contribute to the dataset,
is there a mechanism for them to do so?} 
Yes anybody may open Pull Request with desired changes on GitHub.

\subsection{Limitations and Societal Impact} 
Since our benchmark only contains text in the English language and primarily from commercial domains, its conclusions will only hold for particular types of applications. 
To ensure similar advancements for text/tabular data with low-resource languages \citesi{kann2019towards,lakew2020low,hedderich2020survey}, we encourage the development of a similar benchmark with non-English text. 
We also caution that analysis of text fields may raise privacy concerns as such fields 
may expose arbitrary personal information \citesi{carlini2020extracting,fernandes2019generalised}. Since text fields may contain arbitrary information, they are also prone to introducing spurious correlations in training data that may harm accuracy during deployment \citesi{tu2020empirical} and may be undesirably coupled to protected attributes such as race, gender, or socioeconomic status \citesi{williams2018algorithms}. Basing automated business decisions on customer-generated text could also be more susceptible to adversarial manipulation \citesi{li2019textbugger} than tabular features that customers cannot as easily control.

\clearpage
\bibliographysi{multimodalautoml}
\bibliographystylesi{abbrvnat}

\end{document}